\documentclass[acmtog]{acmart}

\usepackage{cleveref}

\acmSubmissionID{1890}

\copyrightyear{2025}
\acmYear{2025}
\setcopyright{cc}
\setcctype{by}
\acmConference[SA Conference Papers '25]{SIGGRAPH Asia 2025 Conference Papers}{December 15--18, 2025}{Hong Kong, Hong Kong}
\acmBooktitle{SIGGRAPH Asia 2025 Conference Papers (SA Conference Papers '25), December 15--18, 2025, Hong Kong, Hong Kong}\acmDOI{10.1145/3757377.3763945}
\acmISBN{979-8-4007-2137-3/2025/12}

\begin{document}

\title{\includegraphics[height=2ex]{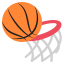}\hspace{1mm}Shoot-Bounce-3D: Single-Shot Occlusion-Aware 3D from Lidar by Decomposing Two-Bounce Light}

\author{Tzofi Klinghoffer}
\email{tzofi@mit.edu}
\affiliation{%
  \institution{Massachusetts Institute of Technology}
  \city{Cambridge}
  \state{Massachusetts}
  \country{USA}
}

\author{Siddharth Somasundaram}
\email{sidsoma@mit.edu}
\affiliation{%
  \institution{Massachusetts Institute of Technology}
  \city{Cambridge}
  \state{Massachusetts}
  \country{USA}
}
\authornote{Both authors contributed equally to this research.}

\author{Xiaoyu Xiang}
\authornotemark[1]
\affiliation{%
  \institution{Meta}
  \city{Menlo Park}
  \state{California}
  \country{USA}
}
\email{xiangxiaoyu@meta.com}

\author{Yuchen Fan}
\affiliation{%
  \institution{Meta}
  \city{Menlo Park}
  \state{California}
  \country{USA}
}
\email{ycfan@meta.com}

\author{Christian Richardt}
\affiliation{%
  \institution{Meta}
  \city{Zurich}
  \country{Switzerland}
}
\email{crichardt@meta.com}

\author{Akshat Dave}
\email{ad74@mit.edu}
\affiliation{%
  \institution{Massachusetts Institute of Technology}
  \city{Cambridge}
  \state{Massachusetts}
  \country{USA}
}

\author{Ramesh Raskar}
\email{raskar@mit.edu}
\affiliation{%
  \institution{Massachusetts Institute of Technology}
  \city{Cambridge}
  \state{Massachusetts}
  \country{USA}
}

\author{Rakesh Ranjan}
\affiliation{%
  \institution{Meta}
  \city{Menlo Park}
  \state{California}
  \country{USA}
}
\email{rakeshr@meta.com}

\renewcommand{\shortauthors}{Klinghoffer et al.}

\begin{abstract}
3D scene reconstruction from a single measurement is challenging, especially in the presence of occluded regions and specular materials, such as mirrors. We address these challenges by leveraging single-photon lidars. These lidars estimate depth from light that is emitted into the scene and reflected directly back to the sensor. However, they can also measure light that bounces multiple times in the scene before reaching the sensor. This \textit{multi-bounce light} contains additional information that can be used to recover dense depth, occluded geometry, and material properties. Prior work with single-photon lidar, however, has only demonstrated these use cases when a laser sequentially illuminates one scene point at a time. We instead focus on the more practical -- and challenging -- scenario of illuminating multiple scene points simultaneously. The complexity of light transport due to the combined effects of multiplexed illumination, two-bounce light, shadows, and specular reflections is challenging to invert analytically. Instead, we propose a data-driven method to invert light transport in single-photon lidar. To enable this approach, we create the first large-scale simulated dataset of \textasciitilde100k lidar transients for indoor scenes. We use this dataset to learn a prior on complex light transport, enabling measured two-bounce light to be decomposed into the constituent contributions from each laser spot. Finally, we experimentally demonstrate how this decomposed light can be used to infer 3D geometry in scenes with occlusions and mirrors from a single measurement. Our code and dataset are released on our \href{https://shoot-bounce-3d.github.io/}{project webpage}.
\end{abstract}


\begin{teaserfigure}
  \includegraphics[width=\textwidth]{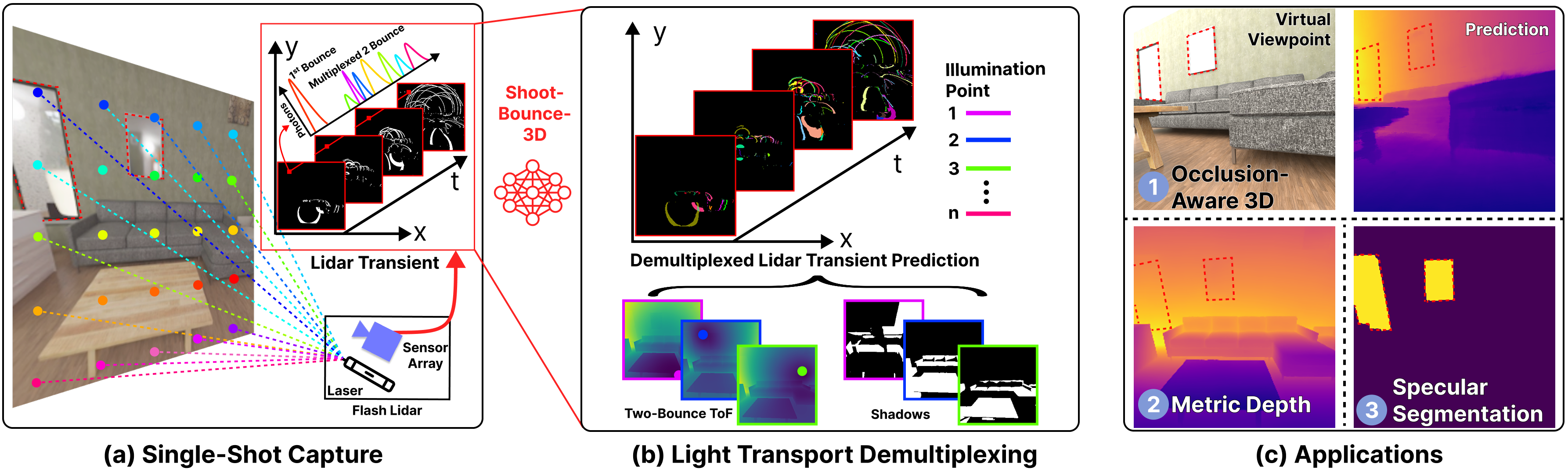}
    \caption{\textbf{Overview.} We introduce \textit{Shoot-Bounce-3D (SB3D)}: a method to decompose temporal light transport in a scene from a single-view, single-shot capture, enabling recovery of 3D geometry, despite specular surfaces and occlusions. \textbf{(a)} A single-photon lidar \underline{\textbf{\textit{shoots}}} light into the scene at multiple points at once, referred to as \textit{multiplexed illumination}. Some light reflects directly back to the sensor, while other light \underline{\textbf{\textit{bounces}}} multiple times first. The lidar captures histograms containing photon intensity over time -- known as \textit{transients}. The multiplexed light mixes together in the transients. \textbf{(b)} We create the first-of-its-kind simulated dataset of multiplexed lidar transients from \textasciitilde100k scenes and use it to train a model to \textit{demultiplex} two-bounce light. \textbf{(c)} Our model enables single-shot \underline{\textbf{\textit{3D}}}, including both dense metric depth and occluded geometry, in the presence of specular surfaces.}
    \label{fig:teaser}
\end{teaserfigure}

\maketitle

\title{Shoot-Bounce-3D: Single-Shot Occlusion-Aware 3D from Lidar by Decomposing Two-Bounce Light}

\section{Introduction}

\label{sec:intro}

Single-shot 3D scene understanding is a long-standing problem in computer vision and graphics -- critical to applications ranging from autonomous vehicles to extended reality. However, recovering 3D information from a single RGB image is ambiguous: lack of multiview correspondences makes metric depth estimation ill-posed, occlusions must be hallucinated, and specular surfaces can be mistaken as holes or ``portals'' in the scene. We present a machine learning (ML) approach to leverage \textit{single-photon lidar} for single-shot 3D reconstruction in scenes with occlusions and specular surfaces.

Single-photon lidars -- composed of a pulsed laser and a single-photon avalanche diode (SPAD) sensor -- shoot light pulses into the scene and measure the time light takes to return to the sensor. Similar to traditional lidar, time taken by the light \textit{directly} reflecting back from the scene -- called time of flight (ToF) -- encodes the depth of illuminated points. However, single-photon lidars can also capture the time taken by light that \textit{indirectly} reflects -- or ``bounces'' -- to other parts of the scene before hitting the sensor. In particular, single-photon lidars measure time-resolved histograms, called \textit{transients}, in which multiple bounces of light appear as multiple peaks (\cref{fig:teaser}a). 

In our work, we focus on two-bounce light: light that has reflected up to two times in the scene. Prior works use two-bounce light in lidar transients to recover dense depth \cite{henley2022bounce}, occluded geometry \cite{klinghoffer2024platonerf}, and material properties \cite{henley2023detection}. However, these works rely on lidars that scan the laser \textit{sequentially} over the scene, one point at a time. Instead, we consider multiplexed illumination -- meaning the scene is illuminated at multiple points \textit{simultaneously}. As a result, measured transients have multiple peaks corresponding to multi-bounce light from \textit{all} illuminated points -- causing prior work to fail. Our investigation of multiplexed illumination is motivated by its use on high-resolution SPADs \cite{henderson20195} found on consumer devices, such as mobile phones, tablets, and headsets \cite{allainiphone,appleLidar2025}.   

Extracting 3D information from transients with multiplexed illumination is challenging due to the ambiguity in mapping \textit{peaks} in the transient to corresponding illumination \textit{points}. In this work, we demonstrate the potential of ML to address this challenge by \textit{demultiplexing} captured transients using data priors (\cref{fig:teaser}b). While ML's application to RGB images has transformed the field of computer vision, single-photon lidar has only recently emerged as a common sensor on consumer devices -- meaning large-scale datasets and ML approaches do not yet exist. Yet, single-photon lidars capture a rich set of features that RGB cameras cannot. We posit that by harnessing these features, ML could enable a new set of abilities in computer vision. Our work is intended as an initial step towards this vision by (1) building the first-of-its-kind simulated multi-bounce transient dataset on \textasciitilde100k scenes (from Aria Synthetic Environments \cite{avetisyan2024scenescript}), and (2) applying it to train our proposed approach, Shoot-Bounce-3D (SB3D), that recovers metric 3D  reconstructions, including in occluded areas and scenes with specular surfaces -- from a single-view, single-shot capture (\cref{fig:teaser}c).

SB3D consists of three steps. First, we estimate the two-bounce time of flight (ToF) of each illumination point. To do this, we train a model to estimate dense depth, which can be directly used to compute the two-bounce ToF for each illumination point. However, not all scene points are illuminated by each illumination point. Scene points that are not illuminated are in shadow; thus, we next train our model to estimate shadow maps for each illumination point. Our model uses the earlier predicted two-bounce ToF for this step. Finally, once both two-bounce ToF and shadows have been predicted for each illumination point, we train an existing method for neural reconstruction to learn 3D scene geometry, including in occluded regions. Because our dataset contains specular surfaces, such as mirrors and windows, our method is robust to these everyday objects. Interestingly, we find that the features learned to demultiplex two-bounce ToF in the first step can also be used to accurately predict specular segmentations in simulation. We posit this is a sign that the features may be a generalizable representation for single-photon lidar. Dataset and code will be released upon acceptance.

\begin{enumerate}
    \item \textbf{Data-Driven Demultiplexing}: From a single-photon lidar measurement of a scene illuminated at multiple points at once, we propose a data-driven method to decompose the two-bounce signal, enabling separation of both two-bounce time of flight and shadows.

    \item \textbf{Occlusion-Aware 3D}: We show that the demultiplexed two-bounce ToF and shadows can be used for single-shot 3D reconstruction, enabling occluded areas to be revealed, despite the presence of specular materials.

    \item \textbf{Large-Scale Multi-Bounce Lidar Dataset:} To enable the above contributions, we introduce a dataset of \textasciitilde100k simulated multi-bounce transient measurements of indoor scenes. This dataset can be used to drive future work in machine learning for single-photon lidar.

    \item \textbf{Generalizable Multi-Bounce Lidar Features}: We find that our demultiplexing model learns features that can be transferred to other tasks, such as specular object segmentation, in simulation. This is a step towards a generalizable multi-bounce transient representation.
\end{enumerate}

\paragraph{Scope of this Work.}
While our work is motivated by the use of high-resolution SPADs \cite{henderson20195, kumagai20217} used with multiplexed point illumination on consumer devices \cite{allainiphone,appleLidar2025}, we acknowledge that these sensors also introduce a variety of practical challenges that are beyond the scope of this work, such as cross talk, hot pixels, blooming, and dead time.
Instead, the purpose of our work is to (a) introduce a data-driven approach to demultiplexing illumination in multi-bounce flash lidars, and (b) show proof-of-concept results.
We do not consider low-resolution SPADs with diffuse illumination \cite{ams_osram_tmf882x,jungerman20223d}.
We assume objects are purely specular or diffuse.
While high-resolution consumer SPADs continue to be developed, they are not yet widely available off the shelf.
As this technology continues to mature, we expect our work to become increasingly relevant.

\begin{figure*}
    \centering
    \includegraphics[width=\textwidth]{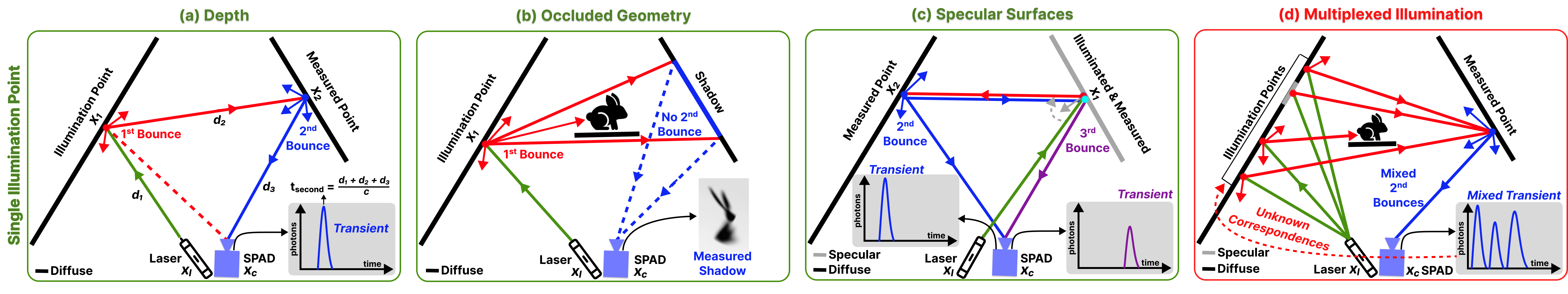}
    \Description{Multi-Bounce Signals.}
    \caption{\textbf{Multi-Bounce Signals.} Shoot-Bounce-3D leverages multi-bounce signals measured from single-photon lidar. Multi-bounce light encodes \textbf{(a)} dense depth (from geometric constraints), \textbf{(b)} occluded geometry (from shadows), and \textbf{(c)} specular surfaces (from two- and three-bounce pairs), but existing techniques assume a single scene point is illuminated at a time, scanning a laser over the scene. However, multi-bounce lidars on consumer devices instead use \textbf{(d)} \textit{multiplexed} illumination, meaning multiple points are illuminated at once -- causing existing methods to fail due to (1) lack of correspondence between two-bounce peaks and illumination points, and (2) mixing of signals from (a), (b), and (c). To resolve these ambiguities, we employ a learning-based technique.}
    \label{fig:flatland}
\end{figure*}

\section{Related Work}

\paragraph{3D from RGB} Recovering 3D information from a single RGB image is ambiguous due to occluded geometry, specular surfaces, and lack of correspondences. Recent foundation models for depth leverage large datasets, learning statistical correlations to address the correspondence ambiguity \cite{Bochkovskii2024, guo2025depth, ke2024repurposing, yang2024depth, yang2024depth2}. Diffusion models are widely used to generate 3D from a single image, including occluded geometry. Several methods generate novel views from an RGB image \cite{liu2023zero, yu2024viewcrafter, sargent2023zeronvs, qian2023magic123, tewari2023diffusion}, though they often focus on objects and struggle in scenes. Starting with \citet{yu2021pixelnerf}, many techniques incorporate data priors in neural radiance fields (NeRF) \cite{mildenhall2021nerf} to learn 3D geometry from single or few images \cite{xu2022sinnerf, gao2024cat3d}. However, NeRF often struggles with the challenge of specular surface materials \cite{verbin2024nerf, tiwary2023orca, ma2024specnerf}. More broadly, specular surfaces cause ambiguities, leading to ``portals'' being hallucinated. Work by \citet{he2021enhanced} and \citet{yang2019my} tries to address this by learning mirror and glass segmentation from RGB. Despite the progress, each of these challenges remains an open problem; rather than relying on RGB, we explore using single-photon lidar for scene-level 3D.

\paragraph{3D from Single-Photon Lidar.} Single-photon lidars offer additional cues for 3D understanding by capturing the distribution of light intensity with travel time, called \textit{transients}. Direct reflections encoded in transients enable photon-efficient depth imaging \cite{shin2016photon,heide2018sub,gupta2019asynchronous}. Recent works also explore multi-view lidar measurements for neural 3D reconstruction \cite{malik2024transient, behari2024blurred, mu2024towards}. Three-bounce information in transients has been extensively used for looking around corners using non-line-of-sight imaging \cite{kirmani2009looking, velten2012recovering, o2018confocal, maeda2019recent, liu2019non, pediredla2019snlos, ahn2019convolutional, shen2024holi}. In this work, we focus on two-bounce light -- which has higher signal quality than three-bounce light due to one less scattering attenuation.  Prior works leverage two-bounce light for dense depth from sparse illumination \cite{henley2022bounce}, seeing behind occluders \cite{henley2020imaging} and specular surface mapping \cite{henley2023detection} -- but require sequentially illuminating the scene one point at a time. Single-photon lidars on consumer devices have multiplexed illumination \cite{appleLidar2025} -- captured transients with mixed light contributions complicate 3D understanding. \citet{lin2024handheld} explore specular surface mapping with multiplexed illumination, but require multiple measurements from different viewpoints. \citet{somasundaram2023role} explore occluded object imaging from multiplexed transients using an analytical approach. We leverage learning to \textit{demultiplex} the captured two-bounce transient to recover depth and occluded 3D geometry in scenes with both diffuse and specular objects in a single shot.

\looseness-1

\paragraph{Data-Driven Methods for Single-Photon Lidar.}

There is a growing interest in applying deep learning methods to single-photon lidar data. Prior works leverage direct bounce information in transients for data-driven depth estimation \cite{lindell2018single, sun2020spadnet,nishimura2020disambiguating,peng2020photon,zang2021non,yang2022deep,plosz2023real}, human pose \cite{ruget2022pixels2pose} and activity recognition \cite{mora2024human}. There are also works on non-line-of-sight imaging from three-bounce transients that develop simulated datasets \cite{chen2020learned} and novel convolutional neural network-based \cite{chen2020learned,sun2020spadnet, mu2022physics, zhu2023compressive,sun2024generalizable,cho2025learning}, transformer-based \cite{li2023nlost,yu2023enhancing} and motion-aware \cite{isogawa2020optical, ye2024plug, chopite2025non} models. To the best of our knowledge, we are the first to bring data-driven methods to two-bounce transients.

\section{Single-Photon Lidar Image Formation Model}

The most common measurement model for lidar only accounts for \textit{direct illumination}:
emitted light that directly reflects back to the sensor. Any other returning light is due to \textit{indirect 
illumination} -- light that interacted with or originated from other parts of the scene -- and is 
treated as ambient noise. However, the secondary light paths from indirect illumination encode useful scene properties, and can be more informative than direct paths. In this section, we show how complex indirect illumination captured by lidar serves as a useful cue for 3D scene geometry, occlusions, and specularity. The ToF dimension is particularly valuable for light transport decomposition when multiple scene points are illuminated simultaneously.

\subsection{Multi-Bounce Light Transport}
\label{subsec:multi_bounce}

\begin{figure*}
    \centering
    \includegraphics[width=\textwidth]{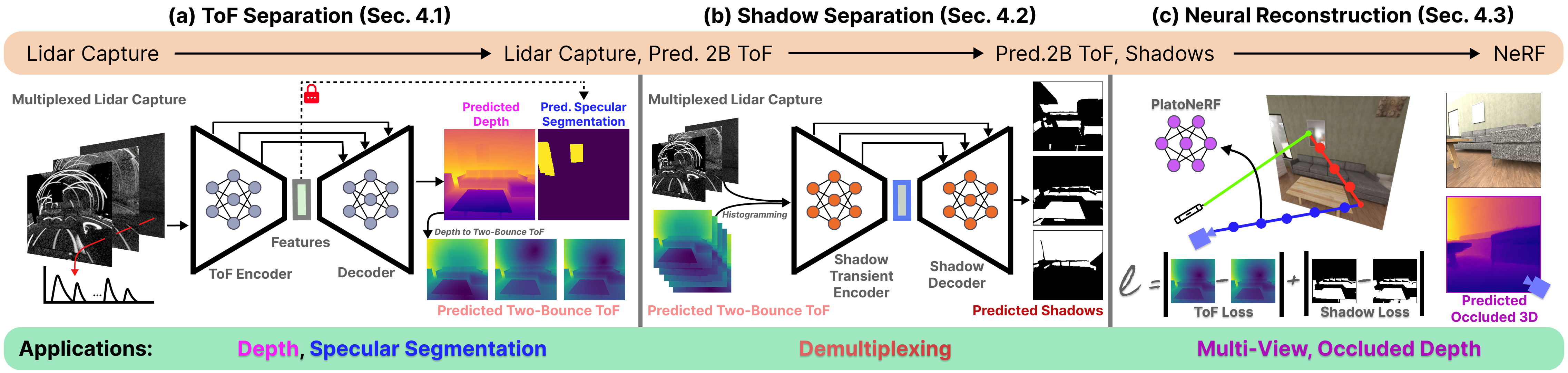}
    \Description{Method overview.}
    \caption{\textbf{Method overview.}
        Shoot-Bounce-3D (SB3D) performs 3D reconstruction from a single lidar measurement.
        The pipeline consists of three steps, each with its own output.
        \textbf{(a)} First, from a measurement taken with multiplexed illumination (meaning multiple points in the scene are illuminated at once), SB3D is trained to learn to predict depth -- allowing the 2-bounce time-of-flight (ToF) for each illumination point to be separated using ray geometries.
        Because our scenes contain specular objects, we find the ToF encoder used for this step also learns features that enable specular object segmentation.
        \textbf{(b)} The predicted 2-bounce ToF is unprojected into histograms and used with the lidar measurement to estimate shadows.
        Using the 2-bounce ToF allows the network to learn \textit{shadow transients}, improving performance.
        \textbf{(c)} Finally, usin g the predicted 2-bounce ToF and shadows, PlatoNeRF can be trained for 3D reconstruction.
    }
    \label{fig:method}
\end{figure*}

Indirect illumination in single-photon lidar arises due to \textit{multi-bounce} light paths. Multi-bounce paths occur when the illuminated scene point becomes a \emph{virtual light source}. A virtual source acts as a light source by reflecting incident light from another source towards other scene points \cite{henley2022bounce}. In this work, a virtual light source can either have isotropic or directional radiance, depending on whether the scene point is diffuse or specular, respectively.

An $n$-bounce path is a light path that consists of exactly $n$ surface reflections before returning to the camera. An example of a multi-bounce path is shown in \cref{fig:flatland}a. A scene point $\mathbf{x}_1$ is illuminated by the laser located at $\mathbf{x}_l$ and imaged by the sensor located at $\mathbf{x}_c$. The light that travels along the path $\mathbf{x}_l \rightarrow \mathbf{x}_1 \rightarrow \mathbf{x}_c$ is referred to as \emph{1-bounce light}. Similarly, light that travels along the path $\mathbf{x}_l \rightarrow \mathbf{x}_1 \rightarrow \mathbf{x}_2 \rightarrow \mathbf{x}_c$ is referred to as \emph{2-bounce} light. The presence, pathlength, and bounce order (i.e., 1-, 2-, or 3-bounce) of multi-bounce light is indicative of the geometry and materials present in a scene. 

\paragraph{Depth.} A key benefit of multi-bounce light is that a scene point doesn't have to be directly illuminated to infer its properties \cite{henley2022bounce, klinghoffer2024platonerf}. Consider the illuminated scene point $\mathbf{x}_1$ and non-illuminated scene point $\mathbf{x}_2$ shown in \cref{fig:flatland}a. The pathlength of the 1-bounce light can be used to infer the 3D position of $\mathbf{x}_1$ using conventional time-of-flight techniques \cite{charbon2014introduction}. Once the location of $\mathbf{x}_1$ is recovered, the location of $\mathbf{x}_2$ can be computed. The light travels a distance of $d_2 + d_3$ along the path $\mathbf{x}_1 \rightarrow \mathbf{x}_2 \rightarrow \mathbf{x}_c$. This distance constrains the possible locations for $\mathbf{x}_2$ to be on the surface of an ellipsoid, with foci at $\mathbf{x}_1$ and $\mathbf{x}_c$ and major axis length $d_1 + d_2$. We also know that $\mathbf{x}_2$ must lie along the pixel viewing direction. Taken together, these two constraints uniquely determine the location of $\mathbf{x}_2$. In this way, 2-bounce light can provide a physical cue for depth even if the scene point isn't directly illuminated.
\looseness-1

\paragraph{Occlusions.} Multi-bounce light can also probe parts of the scene that aren't directly visible to the camera. Consider the example in \cref{fig:flatland}b, where a bunny is behind an obstacle and therefore outside the camera's line of sight. Here, the presence or absence of two-bounce light measured at $\mathbf{x}_2$ is an indication of the presence or absence of an object behind the occluder. If 2-bounce light is absent at $\mathbf{x}_2$ (i.e., $\mathbf{x}_2$ is in shadow), then an object lies along the ray connecting $\mathbf{x}_1$ and $\mathbf{x}_2$. By analyzing the two-bounce intensities (i.e., shadows) along the entire surface on the right wall, the shape of the occluded object can be inferred  \cite{henley2020imaging}. 

\paragraph{Specular Surfaces.} A useful cue to identify specular surfaces is that light returning to the sensor from the specular surface will always arrive after (i.e., have a longer pathlength than) light returning to the sensor from a diffuse surface. For example, consider the case where the virtual source at $\mathbf{x}_1$ is specular, as shown in \cref{fig:flatland}c. In this case, 2-bounce and 3-bounce light will be observed. The 2-bounce light will travel along the path $\mathbf{x}_l \rightarrow \mathbf{x}_1 \rightarrow \mathbf{x}_2 \rightarrow \mathbf{x}_c$, and corresponds to light returning from the diffuse surface. The observed 3-bounce light will travel along the path $\mathbf{x}_l  \rightarrow \mathbf{x}_1 \rightarrow \mathbf{x}_2 \rightarrow \mathbf{x}_1 \rightarrow \mathbf{x}_c$, due to the geometry of specular geometry, and corresponds to light returning from the specular surface. The three-bounce pathlength will be longer than the two-bounce pathlength due to the triangle inequality theorem. A similar argument can be made for the case that $\mathbf{x}_1$ is diffuse and $\mathbf{x}_2$ is specular, in which case 1-bounce light would be observed from the diffuse surface and 2-bounce light would be observed from the specular surface. The resulting property provides a natural cue for detecting the presence of mirrors in a scene. 

\subsection{ToF-Based Multi-Bounce Path Separation}

ToF cameras, such as lidars, can measure the presence and pathlength of multi-bounce light, and separate different light bounces at a pixel due to their high timing precision (picosecond scale). These lidars are now widely available on consumer devices, making them a promising sensing modality for occlusion-aware 3D.

\paragraph{Single-Photon Lidar.}
A single-photon lidar system consists of a pulsed illumination source and a 2D SPAD array. The SPAD array consists of $n_\text{x} \times n_\text{y}$ pixels, and each pixel captures a temporal histogram of $n_\text{t}$ bins.
The $k$th bin of the histogram contains the number of detected photons in the time interval $[k\Delta, (k+1)\Delta]$, where $\Delta$ is the temporal bin width of the sensor.
The resulting SPAD measurement $\mathbf{i} \in \mathbb{R}^{n_\text{x} \times n_\text{y} \times n_\text{t}}$ is a 3D data cube.
There is also a $n_{\text{x}_\text{l}} \times n_{\text{y}_\text{l}}$ laser spot illumination grid within the field of view of the camera, where $n_{\text{x}_\text{l}} \ll n_\text{x}$ and $n_{\text{y}_\text{l}} \ll n_\text{y}$.
Prior works have shown that 3D geometry and specular surfaces can be recovered from $\mathbf{i}$ when each laser spot is illuminated sequentially \cite{henley2022bounce,  henley2023detection}.

\paragraph{Multiplexed Illumination.} In practice, single-photon lidars found on consumer devices typically do not illuminate one laser source at a time: they emit all laser spots simultaneously as shown in \cref{fig:flatland}d. This multiplexed illumination produces an ambiguous signal that integrates the contributions of all laser spots into a single measurement. To recover scene properties, as discussed in \cref{subsec:multi_bounce}, we must ``unmix'' the signal contributions from each light source.

\paragraph{Practical Challenges.} Demultiplexing the contributions of each virtual source is challenging because many of the multi-bounce paths will produce similar pathlengths, making inversion highly ill-posed. Analytical methods for demultiplexing based on 
linear models \cite{somasundaram2023role} and heuristic algorithms \cite{lin2024handheld, henley2023detection} are based on simplifications of the 
underlying light transport that do not generalize to natural scenes. Our key insight is to use a data-driven approach that can learn features directly 
from the physically-constrained, but difficult to model, cues present in transient measurements.

\section{Data-Driven Demultiplexing from Two-Bounce Lidar}
\label{sec:method}

The key challenge that we must solve to leverage multi-bounce light transport in single-photon lidar is demultiplexing illumination. Each laser spot will result in a complex mixture of shadows, specular reflections, and multi-bounce reflections. The presence of multiple laser spots during illumination will further complicate the light transport during capture. The demultiplexing problem entails determining the light transport caused by each individual laser spot.  

In practice, directly reconstructing the per-laser-spot transient is challenging due to its high dimensionality. Instead, we break the problem into steps based on the intuitions in \cref{subsec:multi_bounce}. First, we predict the depth of the visible scene using a neural network (\cref{sec:decompose_tof}). From predicted depth, we estimate separate 2-bounce ToF for each laser spot by tracing the distances $\mathbf{x}_l \rightarrow \mathbf{x}_i \rightarrow \mathbf{x}_{u, v} \rightarrow \mathbf{x}_c$ for all laser spots $\mathbf{x}_i$ and scene points $\mathbf{x}_{u, v}$. For this step, we assume $\mathbf{x}_i$ is known from 1-bounce light. Second, we combine the predicted 2-bounce ToF with the multiplexed measurement to estimate the individual shadows caused by each laser spot (\cref{sec:decompose_tof}). Finally, we combine the 2-bounce ToF and shadows to reconstruct the 3D scene. Demultiplexing is performed in a data-driven manner, and final reconstruction is performed via neural rendering. This pipeline is shown in \cref{fig:method} and implementation details are in the supplement.

\begin{figure}
    \centering
    \includegraphics[width=\linewidth]{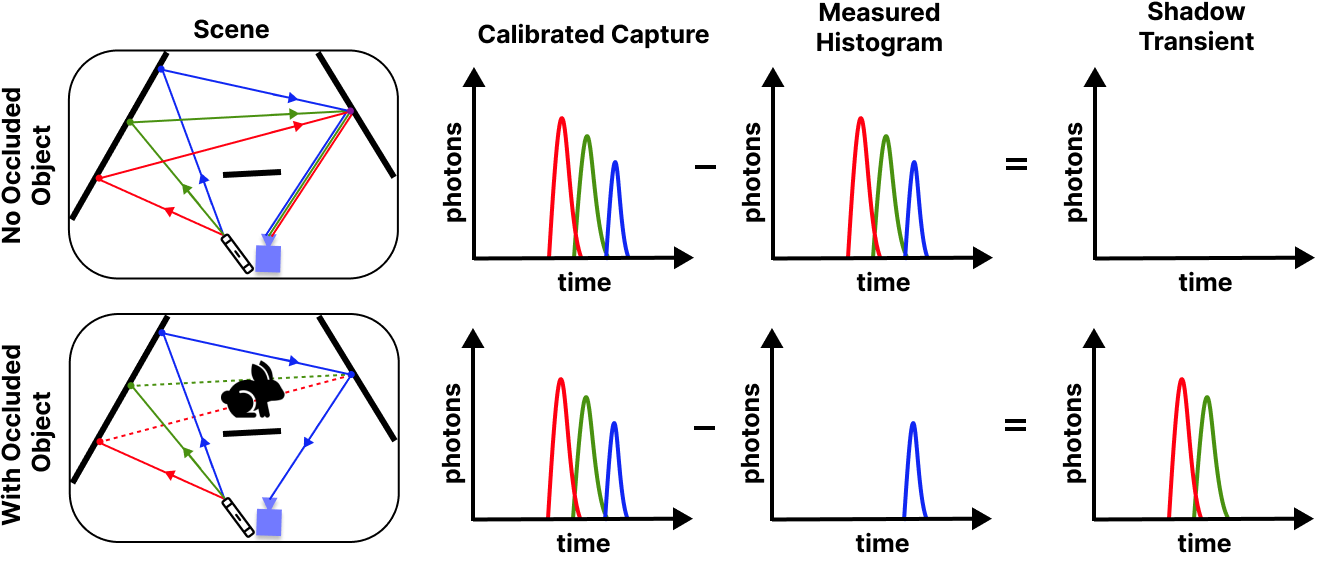}
    \caption{\textbf{Shadow Transients.} We leverage the idea of shadow transients to improve network training for shadow demultiplexing. The key idea is to estimate the light that never reached the sensor due to the object casting a shadow. In the top row, there's no object, so the shadow transient is empty. In the bottom row, two of the three light paths are blocked, so only one peak shows up in the measurement. The shadow transient, on the other hand, measures the two light sources that were blocked by the occluded object. In practice, the calibrated capture is estimated from the data, not measured. As a result, the calibrated capture is input to the network with the measured histogram to prevent errors due to inaccurate shadow transient estimation.}
    \label{fig:shadow_transient}
\end{figure}

\subsection{Demultiplexing Two-Bounce Time-of-Flight}
\label{sec:decompose_tof}

\begin{figure*}
    \centering
    \includegraphics[width=0.95\textwidth]{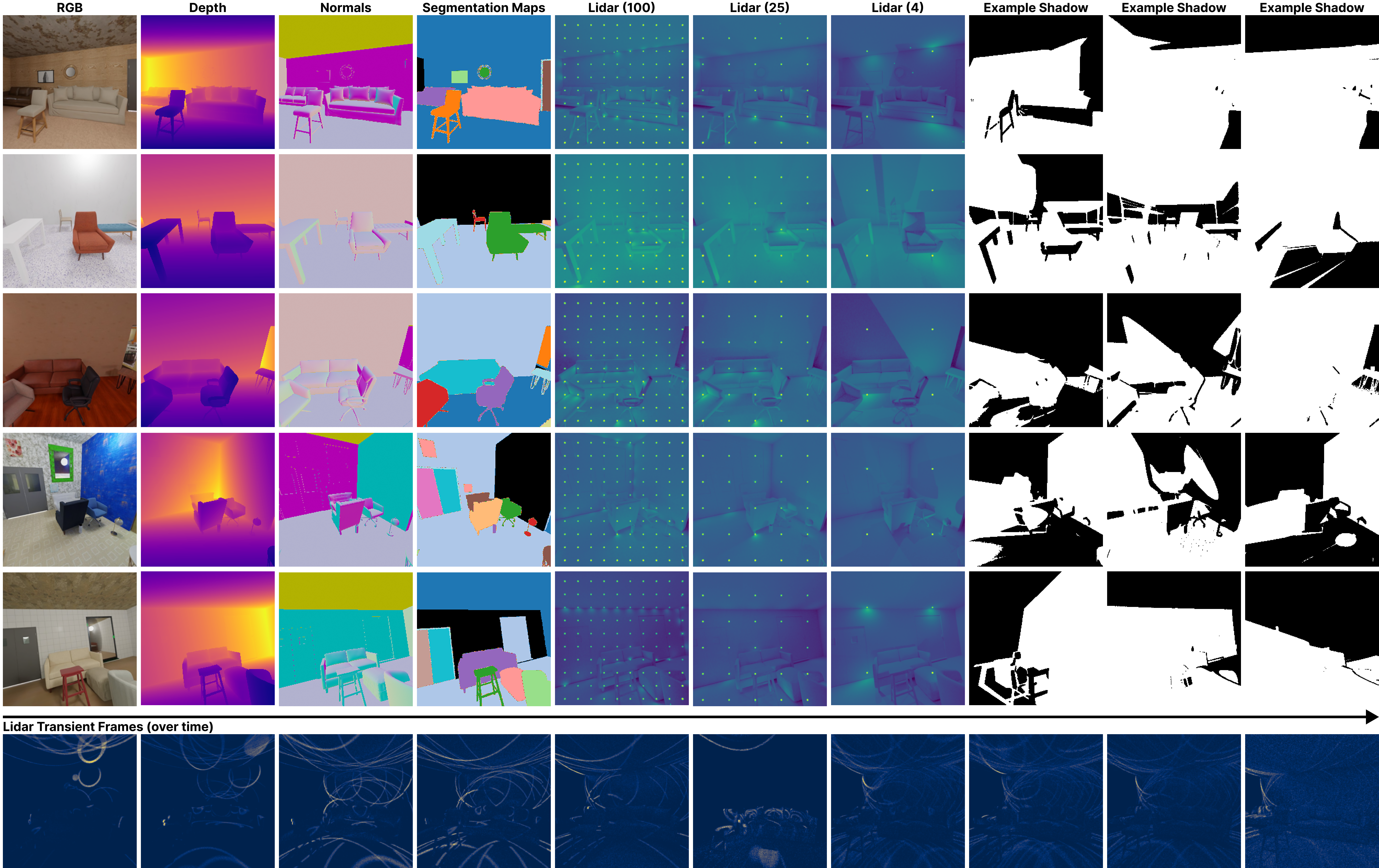}
    \caption{\textbf{Proposed Dataset.} Samples from our simulated dataset of multi-bounce transients for \textasciitilde100k scenes. Our dataset also contains RGB, depth, normals, and segmentation maps for each scene. Transients are simulated with varying amounts of multiplexed illumination -- shown as the intensity maps. Binary shadow maps are provided for each illumination point. The last row shows frames of the simulated lidar transient for an example scene.
    }
    \label{fig:dataset}
\end{figure*}

The 2-bounce ToF of light from individual laser spots is useful for obtaining the visible scene geometry. Predicting the ToF of each laser spot individually, however, is challenging due to the high dimensionality of the output data structure ($n_\text{x} \times n_\text{y} \times n_{\text{x}_\text{l}}n_{\text{y}_\text{l}}$). Instead, we use depth estimation as a proxy task. The depth is a lower-dimensional quantity ($n_\text{x} \times n_\text{y}$) and can subsequently be used to compute the 2-bounce ToF for each laser spot. We train an encoder-decoder to directly learn depth from $\mathbf{i}$ in a supervised fashion. The loss for the depth network consists of a data fidelity and an edge-aware smoothness regularization term:

\begin{equation}
    \mathcal{L}_\text{depth} = \mathcal{L}_\text{data} + \mathcal{L}_\text{smooth} \text{.}
\end{equation}

\noindent Following prior work in depth estimation \cite{godard2017unsupervised, godard2019digging}, the data fidelity term consists of a weighted combination of SSIM \cite{wang2004image} and an L1 loss 

\begin{align}
    \mathcal{L}_\text{data} = \alpha(1 - \text{SSIM}(d, \hat{d})) + (1 - \alpha)\vert d - \hat{d} \vert_1 \text{,}
\end{align}

where $d$ is the ground-truth depth, $\hat{d}$ is the predicted depth, and $\alpha$ is a weight hyperparameter. We set $\alpha{=}0.15$ as in past work \cite{godard2017unsupervised}. We also find that an edge-aware smoothness loss provides a slight improvement in accuracy. This loss encourages the predicted depth to be smooth without blurring edges \cite{godard2017unsupervised}. We use the time-integrated transient measurement $I_{u,v} \!=\! \sum_{t}\mathbf{i}(u, v, t)$ to obtain an estimate of the edges because we don't have access to an RGB image. The resulting loss is 

\begin{align}
    \mathcal{L}_\text{smooth} = \frac{\beta}{N}\sum_{u, v}|\partial_x\hat{d}_{u,v}| \mathrm{e}^{-|\partial_xI_{u,v}|} + |\partial_y\hat{d}_{u,v}| \mathrm{e}^{-|\partial_yI_{u,v}|} \text{,}
\end{align}

where $\hat{d}_{u,v}=\hat{d}(u,v)$, $\partial_x$ and $\partial_y$ are the gradients of the depth and intensity images, and $\beta$ is a hyperparameter which we set to $10^{-3}$.

From the predicted depth, we compute the 2-bounce time-of-flight for each illumination spot, assuming \textit{no occlusions}

\begin{equation}
\label{eq:2b_tof}
    t_{2B}(\mathbf{x}_i, \mathbf{x}_{u,v}; \mathbf{x}_l, \mathbf{x}_c) = \frac{\vert \mathbf{x}_i - \mathbf{x}_l \vert + \vert \mathbf{x}_{u,v} - \mathbf{x}_i \vert + \vert \mathbf{x}_{u,v} - \mathbf{x}_c \vert}{\mathrm{c}} \text{,}
\end{equation}
where $\mathbf{x}_i \in \mathbb{R}^3$ is the location of the $i$th virtual source (known from 1-bounce) and $\mathbf{x}_{u, v} \in \mathbb{R}^3$ is the location of the scene point imaged by pixel $(u, v)$, which can be computed via depth unprojection.

\subsection{Demultiplexing Shadows}
\label{sec:decompose_shadows}

We use the shadows cast by the hidden objects from each illumination spot as the cue for occlusions. Therefore, we aim to recover a set of binary shadow masks $\{\mathbf{s}_1, ..., \mathbf{s}_{n_{\text{l}_\text{x}} n_{\text{l}_\text{y}}}\}$, where $\mathbf{s}_j \in \mathbb{R}^{n_\text{x} \times n_\text{y}}$, \textit{for each illumination spot} from the multiplexed measurement $\mathbf{i}$.

We find that directly training a model to learn $n_{\text{l}_\text{x}} n_{\text{l}_\text{y}}$ shadow masks from the multiplexed measurements, similar to the approach in \cref{sec:decompose_tof}, does not work well. Conditioning the input on the laser spot index similarly does not work well. One possible explanation for poor performance with this approach is due to the inconsistency between the network input and output. The input multiplexed measurements measures the \textit{net amount of light} arriving to the sensor from all illumination spots. The output shadow masks, on the other hand, predicts the \emph{absence of light} from each illumination spot. 

To handle this misalignment, we modify our input based on the concept of \emph{shadow transients} \cite{somasundaram2023role}. The shadow transient $\mathbf{i}_\text{shadow}$ was used in prior work to linearize the forward model for 2-bounce occluded imaging \cite{somasundaram2023role}, and helps align the network input and output in our case. Shadow transients measure the multi-bounce light that \textit{doesn't reach} the sensor due to obstructions from the occluded objects. The shadow transient is $\mathbf{i}_\text{shadow} = \mathbf{i}_\text{calib} - \mathbf{i}$, where $\mathbf{i}_\text{calib}$ is a calibrated capture. This calibrated capture is performed by removing all occluded objects in a scene (leaving only objects that are directly visible to the camera), then capturing the transients, as shown in \cref{fig:shadow_transient}. Rather than removing all occluded objects, $\mathbf{i}_\text{calib}$ can be estimated as

\begin{equation}
    \hat{\mathbf{i}}_\text{calib}(u, v, t) =
    \sum_{i=1}^{n_{\text{x}_\text{l}}n_{\text{y}_\text{l}}}
        A_{i}(u,v) \cdot \delta (t - t_{2B}(\mathbf{x}_i, \mathbf{x}_{u,v}; \mathbf{x}_l, \mathbf{x}_c)) \text{,}
    \label{eq:render_hists}
\end{equation}

where $A_i(u,v)$ is the intensity of two-bounce light returning to pixel $(u, v)$ from virtual source $i$ and $\delta$ is the Dirac delta. We estimate $\hat{\mathbf{i}}_\text{calib}$ using the predicted two-bounce ToF from \cref{sec:decompose_tof}. However, $A_i$ is unknown, resulting in inaccurate estimates of the shadow transients when subtracting $\mathbf{i}$ from $\hat{\mathbf{i}}_\text{calib}$. While past work required hyperparameter tuning for $A_i$, we instead set $A_i=1$ and input both $\mathbf{i}$ and $\hat{\mathbf{i}}_\text{calib}$ to the network. We find that concatenating $\hat{\mathbf{i}}_\text{calib}$ to the input $\mathbf{i}$ significantly improves predicted shadow mask quality. We use a binary cross entropy loss to train the network.

\subsection{Single-Shot 3D Reconstruction}

We use the predicted 2-bounce ToF from \cref{sec:decompose_tof} and the predicted shadows from \cref{sec:decompose_shadows} to train PlatoNeRF \cite{klinghoffer2024platonerf}, a neural reconstruction model to learn 3D geometry. PlatoNeRF requires a separate 2-bounce ToF map and shadow mask for each laser spot. By using the output of our SB3D, PlatoNeRF can be trained without modification, yielding the ability to render depth from extreme novel views that reveal occluded geometry. Because the contributions from all laser spots were captured simultaneously through multiplexed illumination, SB3D enables single-shot capture of the light transport needed for 3D reconstruction. Please refer to the supplement for a comprehensive review of PlatoNeRF.

\subsection{Single-Photon Lidar Feature Generalization}

We take inspiration from existing work in represention learning for RGB images \cite{tian2020contrastive,kolesnikov2019revisiting} and apply it to the context of single-photon lidar. We observe that the features learned from demultiplexing ToF in \cref{sec:decompose_tof} can also be used to perform other tasks, such as specular surface segmentation. To do so, we freeze the pre-trained encoder and train a randomly initialized decoder to predict binary segmentation masks from the learned features. The specular segmentation decoder is supervised with ground-truth segmentation masks and a binary cross-entropy loss.

\begin{figure}
    \centering
    \includegraphics[width=0.8\linewidth]{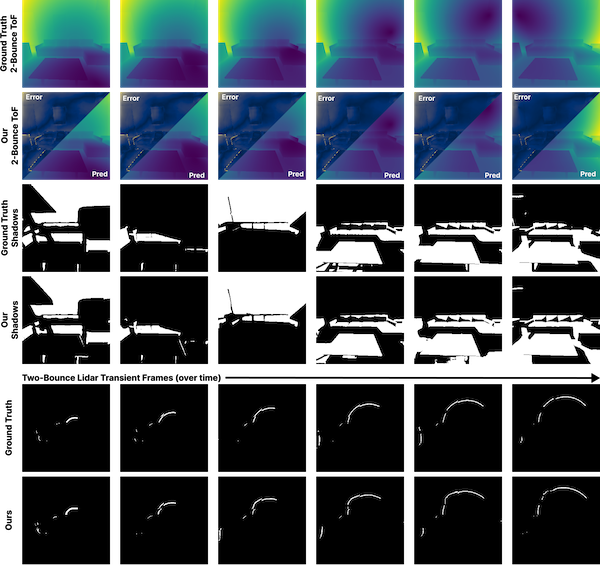}
    \caption{\textbf{Demultiplexing Results.} We show qualitative results for demultiplexing both time of flight (ToF) and shadows. Each column denotes a different illumination source -- our method extracts the two-bounce ToF and shadow maps for each from the multiplexed lidar measurement. The last row shows frames from the predicted ``light in flight'' video (video provided on our \href{https://shoot-bounce-3d.github.io/}{project webpage}), which is rendered by combining the predicted two-bounce ToF and shadows into a transient measurement, allowing visualization of two-bounce light propagation per illumination point.
    }
    \label{fig:demultiplexing}
\end{figure}

\section{Shoot-Bounce-3D Transient Dataset}

\label{sec:dataset}

One of our contributions is a large-scale dataset of simulated lidar transients (\cref{fig:dataset}), built on top of the Aria Synthetic Environments (ASE) dataset \cite{avetisyan2024scenescript}. While there are many point-cloud lidar datasets, only limited single-photon lidar datasets exist. Our dataset contains transients capturing multi-bounce light transport (1st, 2nd, 3rd, etc bounces) that can be exploited for a diverse range of tasks. To the best of our knowledge, this is the first large-scale transient dataset, with past datasets containing \textasciitilde5,000 simulated transients \cite{gutierrez2021itof2dtof}. We render one 256$\times$256 transient at 128 ps temporal resolution for each of the 97,432 ASE scenes (assembled from \textasciitilde8,000 unique objects) we used. These ASE scenes were procedurally created with SceneScript \cite{avetisyan2024scenescript}, which was shown to produce scene geometry sufficiently realistic for real-world generalization. Renderings include single-photon lidar (4, 25, and 100 illumination points), RGB, depth, normals, specular segmentation, instance segmentation, and binary shadow maps. We rendered all data at the same poses as used in the ASE dataset, allowing our dataset to be used in conjunction with ASE in the future. We leverage the lidar transients, ground-truth depth, specular segmentation masks, and shadow masks in our work. More details are available in the supplement.

\section{Experiments}

We present simulated results for demultiplexing, depth estimation, specular segmentation, and occlusion-aware 3D using data containing 25 illumination points. We use $\sim$87k samples for training and 6k for test metrics. We end with proof-of-concept real-world results.

\begin{table}
\caption{\label{table:main_results}
\textbf{Qualitative Results.} We report metrics for each task. For depth and specular segmentation, metrics are computed over 6k test samples. For 3D reconstruction, metrics are computed for predicted multi-view depth, averaged over four scenes (shown in \cref{fig:results}) with 80 novel test views each.}
\vspace{-4mm}
\footnotesize
\centering
\resizebox{0.6\columnwidth}{!}{
\begin{tabular}[t]{lcc}
\multicolumn{3}{c}{\textbf{(a) Depth Estimation}}\\
\toprule
Approach & MAE $\downarrow$ & F1 Boundary $\uparrow$ \\
\midrule
   Bounce Flash Lidar & 0.4922 & 0.0138 \\
   CompletionFormer & 0.4394 & 0.0066 \\
   Depth Anything V2 & 0.1640 & 0.1999 \\
   Depth Pro & 0.1089 & 0.2930 \\
   \bf Shoot-Bounce-3D & \bf 0.0228 & \bf 0.6238 \\
\bottomrule
\end{tabular}
}\\[1mm]

\resizebox{0.6\columnwidth}{!}{
\begin{tabular}[t]{lcc}
\multicolumn{3}{c}{\textbf{(b) Specular Segmentation}}\\
\toprule
Approach & Pixel MAE $\downarrow$ & IoU (\%) $\uparrow$ \\
\midrule
   EBLNet & 0.0117 & 81.21 \\
   \bf Shoot-Bounce-3D & \bf 0.0010 & \bf 86.52 \\
\bottomrule
\end{tabular}
}\\[1mm]

\resizebox{0.6\columnwidth}{!}{
\begin{tabular}[t]{lcc}
\multicolumn{3}{c}{\textbf{(c) Occlusion-Aware 3D Reconstruction}}\\
\toprule
Approach & MAE $\downarrow$ & F1 Boundary $\uparrow$ \\
\midrule
   ZeroNVS & 0.5619 & 0.0090 \\ 
   \bf Shoot-Bounce-3D & \bf 0.0983 & \bf 0.2725 \\
\midrule
   PlatoNeRF Oracle & 0.0950 & 0.3317 \\
\bottomrule
\end{tabular}
}
\end{table}

\subsection{Demultiplexing Results}

First, we investigate the ability of our model to decompose two-bounce light into separated two-bounce ToF and shadows per illumination point. Qualitative results are shown in \cref{fig:demultiplexing} for (a) demultiplexing ToF, (b) demultiplexing shadows, and (c) re-rendering two-bounce transients, which can be visualized as ``light in flight'' videos \cite{velten2013femto}. The transient video for the $i$th laser spot can be rendered by applying the summand in  \cref{eq:render_hists} and setting $A_i(u, v) = \mathbf{s}_i(u, v)$. The last result enables visualization of light propagation per illumination point, as shown in the supp. video. The mean absolute test error for two-bounce pathlength was 0.2736\,m. While correlated to depth error, this error is higher due to the longer paths of two-bounce light. The pixel mean absolute error (MAE) and IoU for predicted shadow maps was 0.0214 and 95.3\%, respectively.

\begin{figure*}
    \centering
    \includegraphics[width=\textwidth]{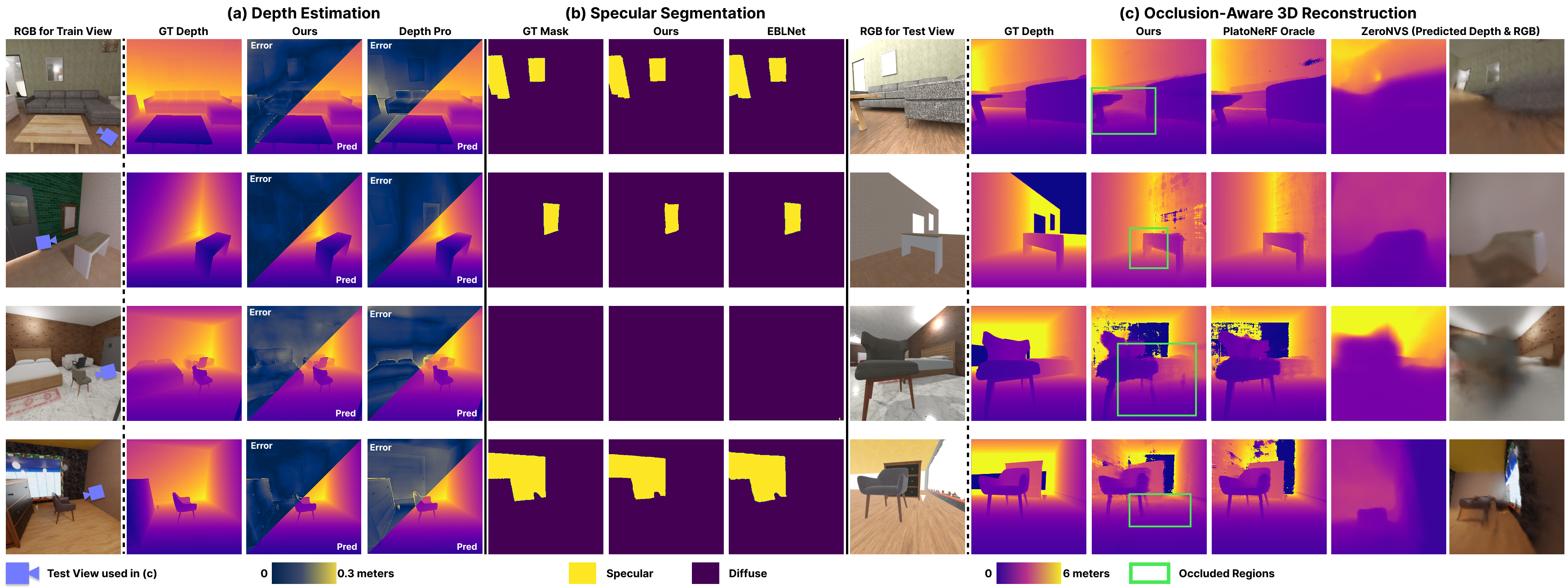}
    \caption{\textbf{Qualitative Results.} We show qualitative results for the tasks of
    \textbf{(a)} depth estimation,
    \textbf{(b)} specular surface segmentation, and
    \textbf{(c)} occlusion-aware 3D reconstruction -- both for our method and the best baseline per task (as well as PlatoNeRF oracle for 3D reconstruction). Our method consistently generates interpretable and accurate results across each of the scenes. For 3D reconstruction, we show novel views that lie in regions occluded from the training view, demonstrating our method's ability to predict demultiplexed shadows that enable inference of hidden geometry. 
    }
    \label{fig:results}
\end{figure*}

\subsection{Depth Results}
\label{sec:depth_results}

Our method is able to predict accurate dense depth from a single image. While dense depth can also be recovered by using a lidar with diffuse illumination and dense pixel array, existing consumer devices such as the iPhone use point illumination \cite{appleLidar2025,allainiphone}, which results in sparse depth but better range. Our method, which also uses point illumination, also recovers dense depth due to its use of both one-bounce and two-bounce paths.

\paragraph{Baselines.} We compare to a physics-based lidar method, a learned RGB/lidar fusion method, and RGB depth foundation models:

\begin{enumerate}
    \item \textbf{Bounce-Flash (BF) Lidar} \cite{henley2022bounce} uses geometric constraints to estimate depth from two-bounce lidar. Because BF lidar assumes scanned illumination, we adapt it to the multiplexed setting by using it to compute depth candidates for all (two-bounce peak, illumination point) pairs, via peak finding. Intuitively, if a scene point is not in shadow for two illumination points, then BF Lidar will yield approximately the same depth for both. We compute the mode of the discretized depth candidates to compute depth.

    \item \textbf{CompletionFormer (CF)} \cite{zhang2023completionformer} recovers dense depth from monocular RGB and sparse lidar. We use RGB images rendered at the same view as our lidar transients and depth from our sparse illumination points (computed from first bounce) as input.

    \item \textbf{Depth Anything V2 \& Depth Pro} \cite{yang2024depth2,Bochkovskii2024} are RGB foundation models for monocular depth, trained on large sets of real and simulated RGB. Although both models can compute metric depth, we found this inaccurate. We rescaled predictions via a least squares regression on 25 anchor points, acquired from the one-bounce returns in our lidar measurements.
\end{enumerate}

\begin{figure*}
    \centering
    \includegraphics[width=0.85\textwidth]
    {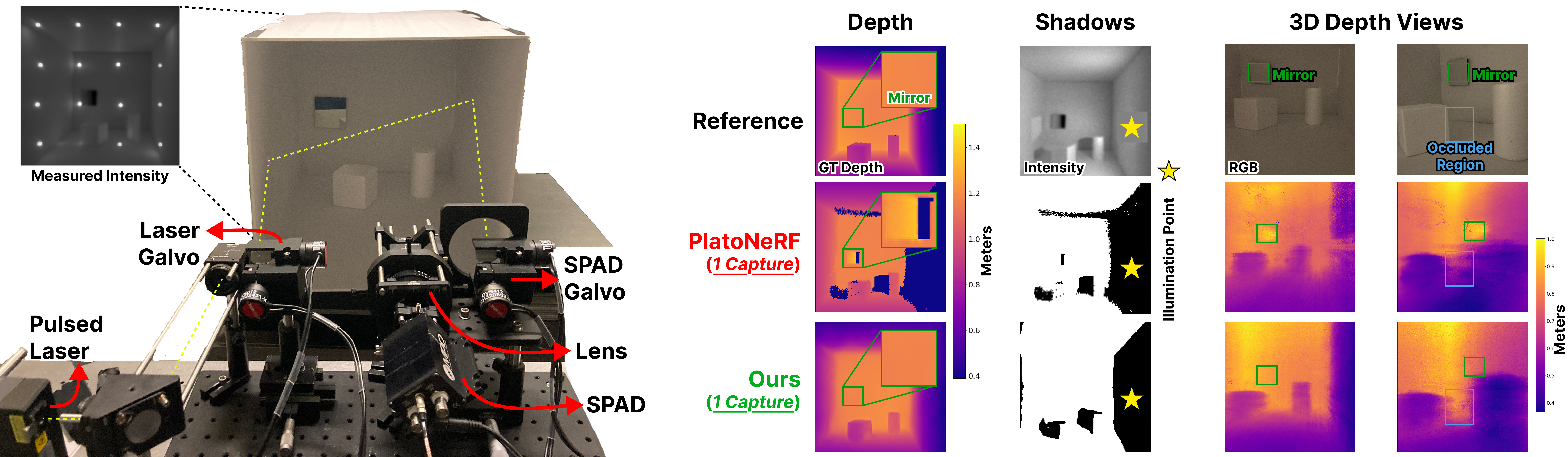}
    \caption{\textbf{Real-World Results.} We provide proof-of-concept real-world results on a new dataset that we capture with multiplexed illumination (16 laser spots). We compare our method to PlatoNeRF for depth, shadows, and 3D depth views. Since our goal is single-shot 3D, we restrict PlatoNeRF to a single capture, which, since it is unable to handle multiplexing, means it is trained with 1 illumination point. Our method, on the other hand, learns to demultiplex 16 illumination points, leading to better performance, especially in specular regions. A comparison to PlatoNeRF with more captures is provided in the supplement.
    }
    \label{fig:real-world}
\end{figure*}

\paragraph{Metrics.} We compute metric depth mean absolute error and scale-invariant boundary F1 defined by \citet{Bochkovskii2024}.

\paragraph{Results.} We find that our method significantly outperforms the baselines. Quantitative results are provided in \cref{table:main_results}a and qualitative results and error maps for our method and Depth Pro are shown in \cref{fig:results}. Depth Pro produces qualitatively similar depths as the ground truth, but struggles to preserve scale, even after rescaling with anchor points. In addition, we notice higher depth error around edges with Depth Pro compared to our method. Because BF Lidar has no learnable mechanism for denoising or infilling, it produces noisy depth maps and is unable to resolve depth for shadowed pixels. CF struggles when provided depth from only 25 points. Even when 100 points are provided, it still achieves only 0.149\,m MAE.

\subsection{Specular Surface Segmentation Results}

\paragraph{Baselines.} We compare our method for specular segmentation to EBLNet \cite{he2021enhanced}, which learns to segment glass and mirrors from an RGB image. For fair comparison, we retrain EBLNet on our dataset. We considered lidar comparisons, but found that existing methods either do not focus on multiplexed illumination \cite{henley2023detection} or use a different hardware setup \cite{lin2024handheld}.

\paragraph{Metrics.} Specular surface segmentation is a binary segmentation task where values of one indicate a specular surface and values of zero indicate a diffuse surface. We report pixel mean absolute error ($\sum_{u,v}|s_{u,v} - \hat{s}_{u,v}|$, where $s$ and $\hat{s}$ are ground-truth and predicted segmentation values per pixel), and intersection over union (IoU).

\paragraph{Results.}
We find that our method is able to detect mirrors with high accuracy, outperforming EBLNet. We posit the increase in performance is due the availability of physical cues correlated to specular surfaces (\cref{fig:flatland}c), whereas, in RGB images, detecting specular surfaces is inherently ambiguous (a specular surface and a ``portal'' often look identical). While our work focuses on two-bounce signals, since we use the full transient as input to our model, it may also use three-bounce signals. We ablate this further in the supplement.

\subsection{3D Reconstruction Results}

Finally, we evaluate SB3D's 3D reconstruction quality. Recall from \cref{fig:method} that 3D reconstruction is done per-scene by supervising PlatoNeRF with SB3D's predicted two-bounce ToF and shadows.

\paragraph{Baselines.} We compare our method to the following approaches. We also attempted a comparison with \citet{somasundaram2023role}, but the method assumes that a calibrated capture $\mathbf{i}_\text{calib}$ is available, as described in \cref{sec:decompose_shadows}, which is not the case in practice. 

\begin{enumerate}
    \item \textbf{ZeroNVS} \cite{sargent2023zeronvs} trains a NeRF from a single RGB image via score distillation sampling of a diffusion model trained on a large-scale dataset. 
    \item \textbf{PlatoNeRF Oracle} \cite{klinghoffer2024platonerf}: Our method is built atop PlatoNeRF -- a recent method for 3D reconstruction from two-bounce lidar. Therefore, we train an ``oracle'' PlatoNeRF model from ground truth two-bounce ToF and shadows for each scene to disentangle performance of our method and performance of PlatoNeRF.
\end{enumerate}

\paragraph{Metrics.} We compute MAE and boundary F1 on multi-view depth rendered from NeRF as a proxy for 3D reconstruction.

\begin{figure}
    \centering
    \includegraphics[width=0.9\linewidth]{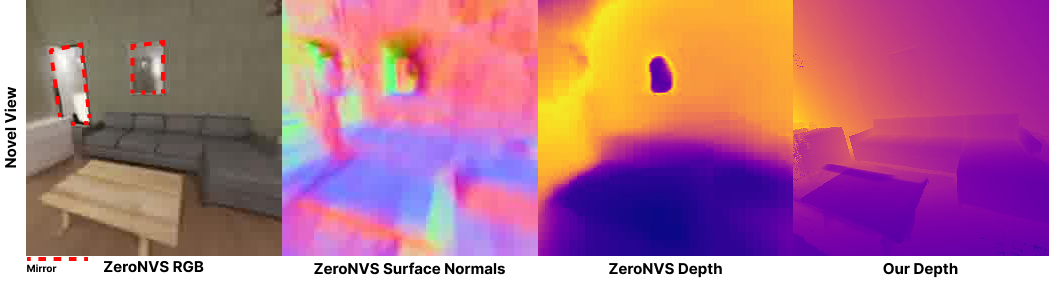}
    \caption{\textbf{Portals in Mirrors.} RGB-based methods -- such as ZeroNVS (shown above) -- are prone to hallucinating ``portals'' inside mirrors due to the lack of cues to distinguish the mirror from a physical space. By leveraging multi-bounce transients, our method can handle specular surfaces, such as mirrors, enabling accurate 3D reconstruction even in the presence of mirrors. 
    }
    \label{fig:mirror_portal}
\end{figure}

\paragraph{Results.} Qualitative results for our method and ZeroNVS are shown in \cref{fig:results}c. We find that our method is able to accurately reconstruct not only visible regions -- but also occluded regions, yielding interpretable and detailed geometry. ZeroNVS is able to discern coarse structure -- and in some cases, such as the last scene, carve out empty space in occluded regions, but fails to recover detailed geometry. Lack of detailed geometry may be because of geometric inconsistencies that emerge from training with a diffusion model. In addition, ZeroNVS relies entirely on hallucination in regions that are fully occluded, whereas SB3D relies on a physically meaningful quantity -- demultiplexed shadows. Although ZeroNVS struggles to perform accurate novel view synthesis in the extreme views and occluded areas emphasized in this work, we note that it performs better with small view changes without occlusion, as shown in \cref{fig:mirror_portal}, yielding better, albeit still less accurate depth. \Cref{fig:mirror_portal} also highlights the importance of understanding specular surfaces during 3D reconstruction, which our method is able to do due to earlier steps implicitly learning about them when demultiplexing.

\subsection{Real-World Results}

We provide an overview of our real-world results and refer the reader to the supplement for more details and discussion.

\paragraph{Dataset.} We collect a real-world dataset by scanning a single-pixel SPAD (MPD PDM Series) over the field of view of the scene (containing a cube and cylinder inside a room with a mirror on the wall) using a two-axis scanning galvanometer (ThorLabs GVS412). This process is repeated for 16 illumination points and the transients are summed to create a multiplexed measurement for testing. The real-world data is $256\times256$ with a temporal resolution of 32\,ps.

\paragraph{Results.} To validate our method, we retrain our models with a simulated dataset of 10k scenes containing a cube, cylinder, and mirror randomly placed inside a room of varying scale. During training, we add Poisson noise and Gaussian timing jitter to the transients. We then test the models on the real-world dataset and find that they are able to recover accurate depth and shadow masks, enabling 3D reconstruction and outperforming PlatoNeRF in the single-shot setting (\cref{fig:real-world}). Our reconstruction is created by manually selecting the 4 best shadow masks, since, as shown in the supplement, some predictions contain artifacts. These artifacts can be mitigated in the future by improving SNR of test data, incorporating more realistic noise in simulated training data, and introducing real-world data in training. Our depth error is 0.028 m and boundary F1 is 0.556. We find that as the number of captures used to train PlatoNeRF increases, so does its performance, but SB3D remains competitive, as shown in the supplement. These results demonstrate that the proposed method can be successfully extended to real-world data.

\subsection{Ablations}

\begin{figure}
    \centering
    \includegraphics[width=0.85\linewidth]{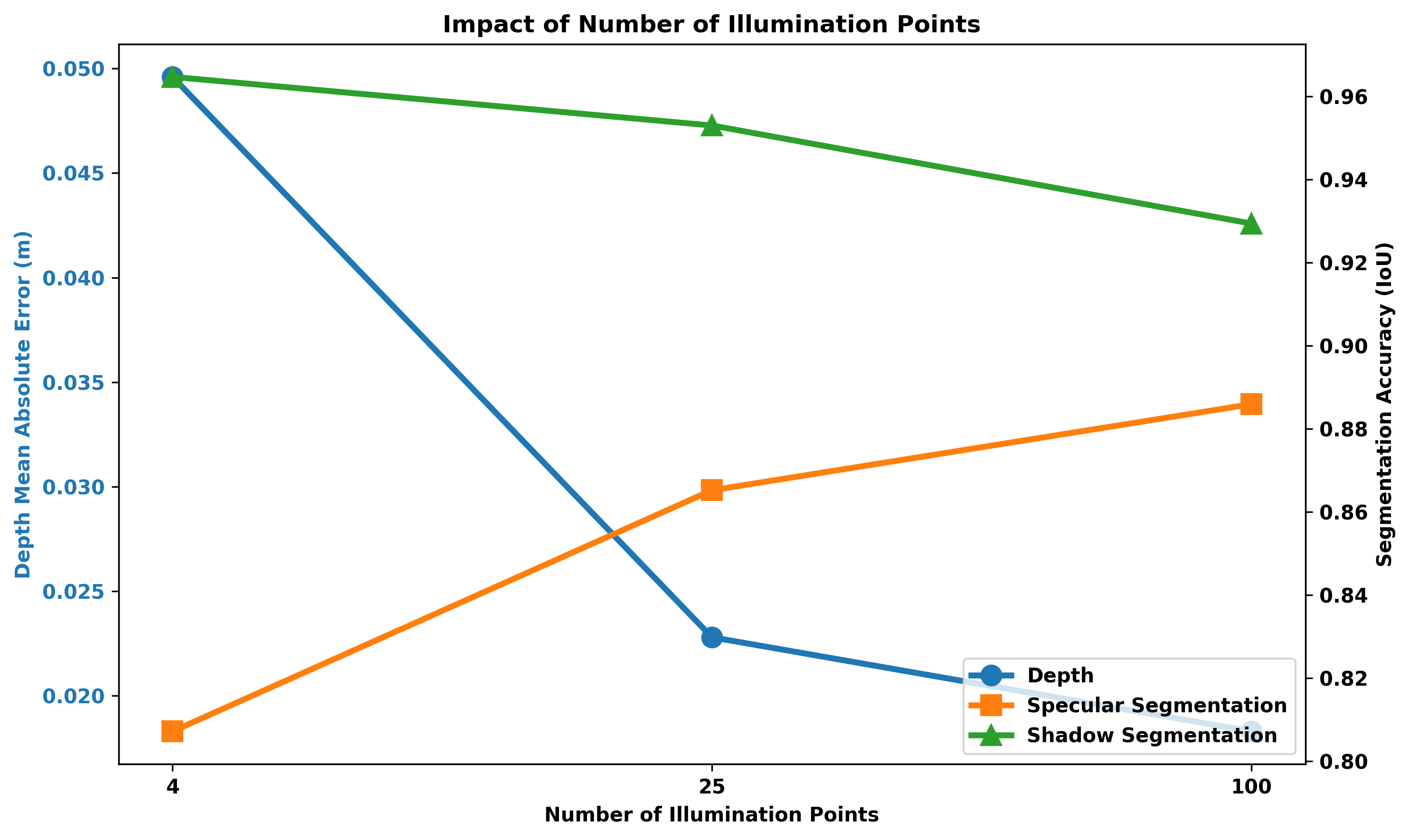}
    \caption{\textbf{Ablation: Number of Illumination Points.} We show the trade-off in performance as the number of illumination points is varied from 4 to 25 to 100. While depth and specular surface estimation improve with more illumination points -- despite increased ambiguities from multiplexing -- shadow mapping becomes less accurate. Therefore, a Pareto optimum exists that balances the accuracy of all three tasks.
    }
    \label{fig:num_illumination_points}
\end{figure}

\begin{figure}
    \centering
    \includegraphics[width=0.9\linewidth]{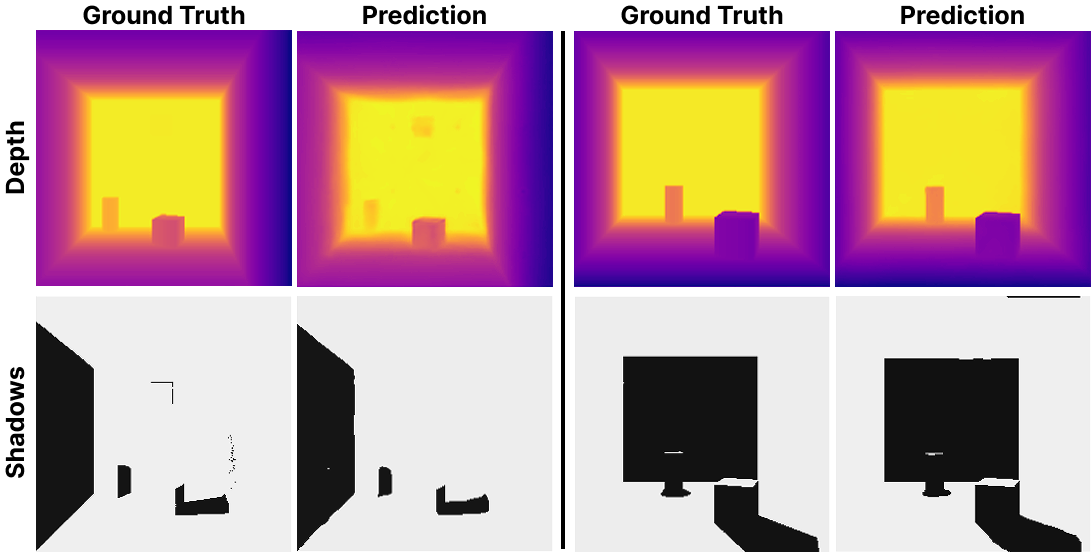}
    \caption{\textbf{Pulse Shape, Noise, \& Timing Jitter.} To demonstrate that the ideas introduced in this work can generalize to realistic pulse shapes, noise, and timing jitter, we (a) convolve our simulations with real-world sensor pulses, (b) add Poisson noise to the histogram intensities (such that 2-bounce peak photon counts range from 10 to 400), and (c) add Gaussian timing jitter to the histogram peaks (50 ps full width at half max). Accurate depth and shadow estimation show robustness to some practical challenges with lidar.
    }
    \label{fig:noise}
\end{figure}

We find our method continues to work when retrained on separate datasets with 4 or 100 illumination points (\cref{fig:num_illumination_points}). We also ablate adding realistic pulse shapes, noise, and timing jitter on a simplified dataset in \cref{fig:noise}. More details on these ablations, as well as additional ablations, including on shadow estimation, training on 1- and 2-bounce light only, amount of training data, out-of-distribution geometry, and temporal resolution, are in the supplement.

\section{Conclusion}

We present a method for single-shot estimation of depth and 3D scene geometry in the presence of specularities using single-photon lidars with \textit{multiplexed} illumination. At the heart of our method is the first large-scale simulated dataset of \textasciitilde100k lidar transients, which enables our learned technique for \textit{demultplexing} ToF and shadows from a single lidar measurement. Not only does demultiplexing enable 3D reconstruction, but we also find that the learned features could generalize to other tasks, such as specular segmentation.

\paragraph{Limitations.}
Due to our pipeline design, errors from ToF separation propagate to shadow separation. In addition, this work does not consider practical challenges that arise on consumer SPADs, such as cross talk, hot pixels, dead time, and blooming. Our work is a step towards enabling single-shot 3D on these sensors. 

\paragraph{Future Work.}
Our dataset and method open opportunities for future work in single-photon lidar foundation models and fusion with RGB images. In addition, future work is needed to automate the selection of top shadow mask predictions and to reduce the propagation of errors, potentially through end-to-end methods.

\begin{acks}
We thank Suvam Patra and Armen Avetisyan for support with Aria Synthetic Environments. Tzofi Klinghoffer is supported by the Department of Defense (DoD) National Defense Science and Engineering Graduate (NDSEG) Fellowship Program. Siddharth Somasundaram is funded by the National Science Foundation (NSF) Graduate Research Fellowship Program (Grant \#2141064). 
\end{acks}

\bibliographystyle{ACM-Reference-Format}
\bibliography{sample-base}

\clearpage
\appendix

\begin{figure*}[t]
    \centering
    \includegraphics[width=0.95\textwidth]{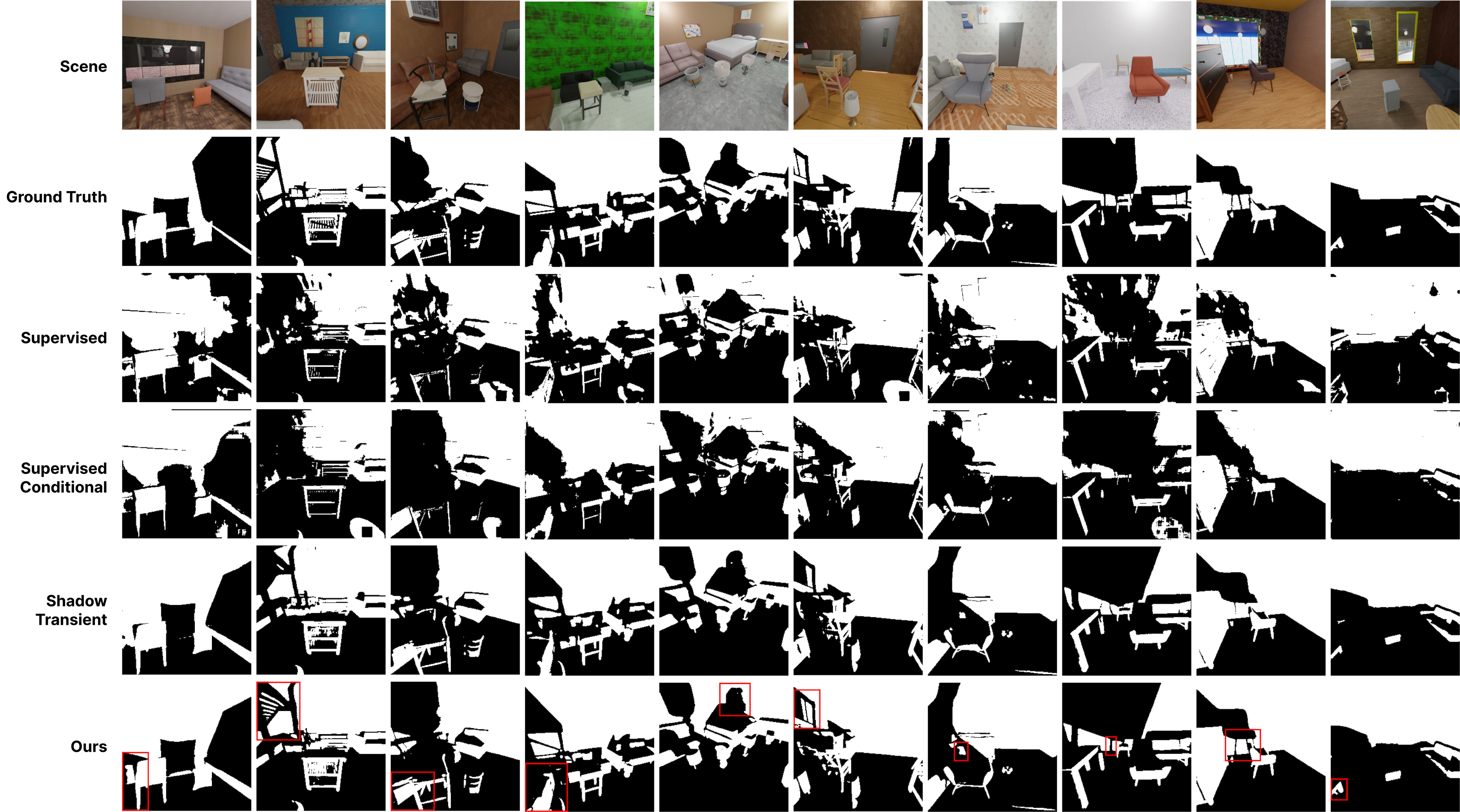}
    \caption{\textbf{Shadow Mapping Approaches Ablation}. We compare different methods for predicting shadow masks from multiplexed lidar transients. All results are for measurements with 25 laser spots. We compare: supervised learning (transient $\rightarrow$ 25 shadows), supervised learning conditioned on laser spot index ((transient, index) $\rightarrow$ shadow), shadow transients (\textit{abs}(2-Bounce ToF transient -- measured transient) $\rightarrow$ shadow), and ours ((2-Bounce ToF transient, measured transient) $\rightarrow$ shadow). Results indicate that the use of shadow transients -- whether explicit or implicit (ours) significantly improves results. Providing both the predicted 2-bounce ToF transient and measured transient as input to the network, rather than explicitly computing the shadow transient, slightly improves detail and performance, as shown by the regions in the red boxes.
    }
    \label{fig:shadow_mapping_approaches}
\end{figure*}

\section{Ablations}

\begin{figure}
    \centering
    \includegraphics[width=1.0\linewidth]{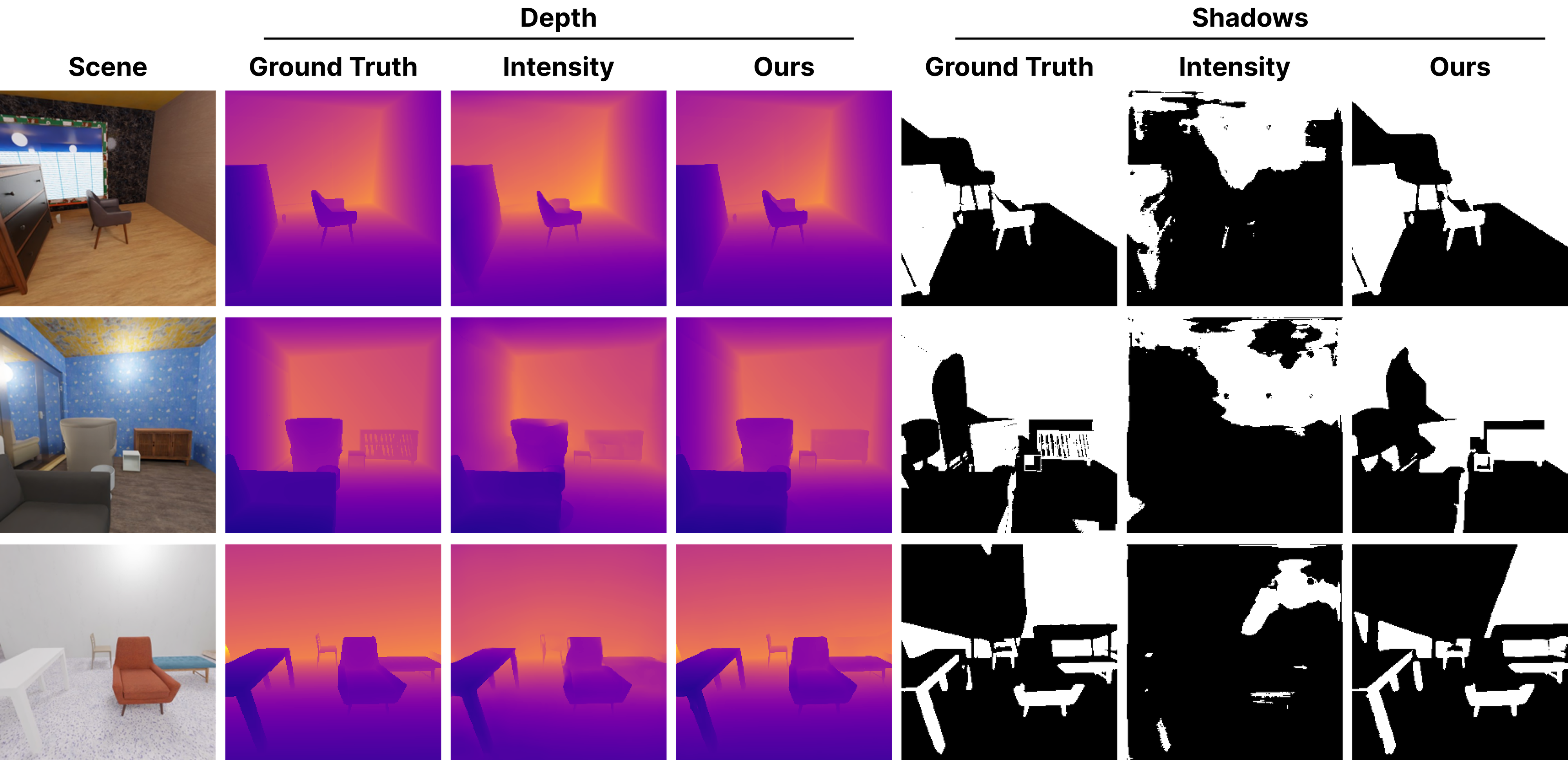}
    \caption{\textbf{Intensity Only Ablation.} We compare depth and shadow estimation when using intensity images from the SPAD (obtained by summing along the temporal dimension) vs the full transient (ours). Using the full transient improves both depth and shadow results. 
    }
    \label{fig:intensity_ablation}
\end{figure}

\begin{figure}
    \centering
    \includegraphics[width=0.7\linewidth]{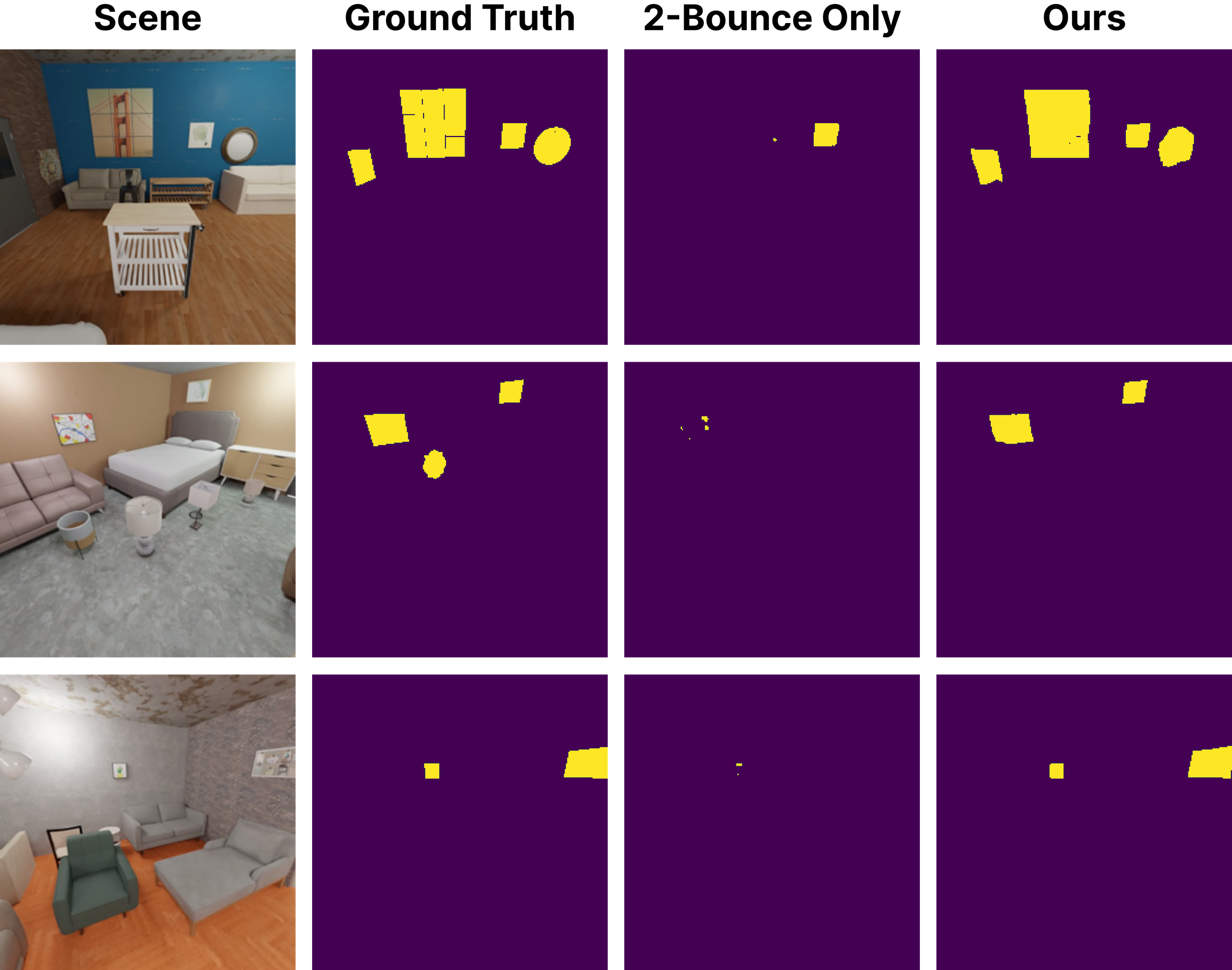}
    \caption{\textbf{Two-Bounce Only Ablation.} We compare specular segmentation for models trained with transients rendered with only 1- and 2-Bounce peaks vs all peaks (ours). We randomly set pictures to be specular and find that only using 2-bounce information is not sufficient -- thus, we posit this model also relies on 3-bounce information.
    }
    \label{fig:2b_ablation}
\end{figure}

We include ablations to better understand the impact of: (1) the proposed use of \textit{shadow transients} for shadow mapping, (2) number of illumination points, (3) single-photon lidar's temporal dimension, (4) light that has bounced \textit{more than two} times, (5) modeling pulse shape, noise, and timing jitter, (6) amount of training data, and (7) out-of-distribution geometry at test time. We include quantitative results for these ablations in text below and qualitative results in \cref{fig:shadow_mapping_approaches}, \cref{fig:intensity_ablation}, and \cref{fig:2b_ablation}.

\label{sec:ablations}

\subsection{Shadow Mapping Approaches}
Our approach draws on the idea of shadow transients to map lidar measurements to binary shadow maps for each illumination source. Rather than only using the multiplexed measurement as input, we also provide predicted two-bounce ToF. To understand the benefit of this approach, we compare it to the naive approach of using only the multiplexed input with supervised learning. Since there is no cue for which shadow to generate given a multiplexed measurement, we try two approaches: (1) predict shadows for all illumination points in a forward pass, (2) condition on illumination point and predict only the corresponding shadow in a forward pass. We find that the model is unable to learn accurate shadows with either of these approaches. While our proposed method -- which uses predicted two-bounce ToF as an additional input -- yields 0.0186 MAE and 0.959 IoU (Fig. \ref{fig:shadow_mapping_approaches}, row 6), (1) yields 0.0984 and 0.788, respectively, and (2) yields 0.0896 and 0.799, respectively. Qualitative results for (1) and (2) are shown in Fig. \ref{fig:shadow_mapping_approaches}, rows 3 and 4, respectively. We also tried explicitly computing the shadow transients by taking the absolute difference of the predicted two-bounce ToF transient and measured transient, and using this as input, which, while still accurate, results in slightly worse performance (0.0231 MAE, 0.950 IoU), as shown in Fig. \ref{fig:shadow_mapping_approaches}, row 5. These results indicate that using shadow transient information -- either implicitly or explicitly -- is critical for accurate shadow mapping from multiplexed lidar measurements.

\subsection{Number of Illumination Points}

We vary the number of illumination points -- using 4, 25, and 100 -- and study its impact on depth estimation, specular surface segmentation, and shadow mapping (a proxy for occluded 3D reconstruction; i.e., inaccurate shadows lead to poor reconstruction). Intuitively, more illumination points increases ambiguity -- as there are more unknown correspondences of peaks to illumination points. However, we find that this only reduces performance for shadow mapping -- more illumination points results in higher performance when estimating either depth or specular surfaces. Results are shown in \cref{fig:num_illumination_points} of the main text. As more illumination points are added, there is more redundancy in depth information -- since each illumination points provides a depth cue for all points light reflects to. Similarly, more illumination points increases the odds of a specular cue becoming available -- since specular cues depend on light from a specular surface eventually being reflected back to the sensor. However, increasing the number of illumination points presents a serious challenge for the model to separate shadows, resulting lower performance. Thus, we hypothesize there is a \textit{pareto optimum} that exists for number of illumination points, which results in higher depth, specular surface, and shadow accuracy. In our work, this optimum occurs at 25 illumination points. More work is needed to understand if the trend of depth improving with more illumination points continues with significantly more illumination points -- on one hand, these additional illumination points provide redundancy in depth information, but, on the other, they increase the peak to illumination point correspondence ambiguities.

\subsection{Intensity Only Measurements}
One of the primary hypotheses of this work is that the temporal dimension of transient measurements contains information that can enable new advancements in 3D computer vision. Thus, for each of the tasks that we perform -- estimation of depth, specular surfaces, and occluded geometry -- we also conduct a baseline using an \textit{intensity} image from a single-photon lidar. The intensity image is simply the sum of the transient along the temporal dimension. As expected, using the intensity image for training and inference on each task results in significantly worse performance than our method, which leverages the full information in the 3D transient. Using the intensity image results in  0.174 m mean absolute error for depth estimation, 0.703 IoU for specular segmentation, and 0.700 IoU for shadow mapping. While a significant drop in performance, interestingly, the intensity image still contains relevant cues for modest performance on each task, though we found the resulting shadow maps are not sufficiently accurate to perform 3D reconstruction.

\subsection{Two-Bounce Only Measurements}
\label{sec:two_bounce_ablation}

We also study the impact of training on transients that only contain first and second bounce information to understand which signals the model has learned to exploit. We do this by re-rendering the transient dataset in MitsubaToF and setting $max\_depth$ to three (whereas the main dataset contains all bounces). We re-train our models for each task, allowing us to understand the importance of three or more bounces of light based on the change in performance per task. Depth MAE increases from 0.0228 to 0.0255\,m ($\Delta$0.0027\,m), shadow mapping IoU increases from 0.954 to 0.964 ($\Delta$0.01), and specular surface IoU drops from 0.865 to 0.767 (-$\Delta$0.098). Thus, while three bounce has a minimal impact on depth and shadows, it has a significant impact on specular surface segmentation. This result matches our intuition that three-bounce signals can contain information about specular surfaces. While two-bounce signals may indicate a specular surface based on the presence of an extra measured peak at scene points that receive light reflected directly off a specular surface, three-bounce signals are empirically more helpful. In particular, we posit that the model has not only learned to exploit diffuse-specular-diffuse light paths, as done in past work, but also diffuse-diffuse-specular light paths, which may be a fruitful direction to investigate in future work. This hypothesis is based on the diffuse-diffuse-specular signal that is visually evident when watching the light-in-flight transient videos.

\subsection{Noise and Timing Jitter} 
\label{sec:noise}
Since our method is trained with simulated measurements, we ablate its ability to work when trained on measurements with realistic pulse shapes, noise, and timing jitter. As done by \citet{chen2020learned}, we follow the protocol established in \citet{hernandez2017computational} for modeling realistic SPAD measurements. First, we convolve the rendered histograms with a pulse measured with a real-world sensor (MPD PDM Series). Next, we add Poisson noise and Gaussian timing jitter to the histograms. We sample the rate from a uniform distribution, leading to 2-bounce peak photon counts ranging from 10 to 400. We add 50 ps timing jitter (FWHM), which corresponds to 6.25 bins at 8 ps resolution. For this ablation, we use a small dataset containing geometric primitives to ease training time. The dataset consists of 10k training samples. Each scene also contains a mirror. We demonstrate both accurate depth estimation (which can then be used to estimate two-bounce ToF) and shadow mapping using our method trained on this dataset. While our method is able to work in these conditions, we found that the 2D U-Net model was significantly less accurate than the ``2.5D'' U-Net alternative proposed in \cref{sec:implementation}. While the ``2.5D'' U-Net was not used in the main experiments due to higher training time, our use of a smaller and simpler dataset in this ablation, allowed us to use it. We posit that the 3D encoder enables the model to learn more robust temporal features, allowing it to generalize better under larger amounts of noise.

\begin{figure}
    \centering
    \includegraphics[width=0.9\linewidth]{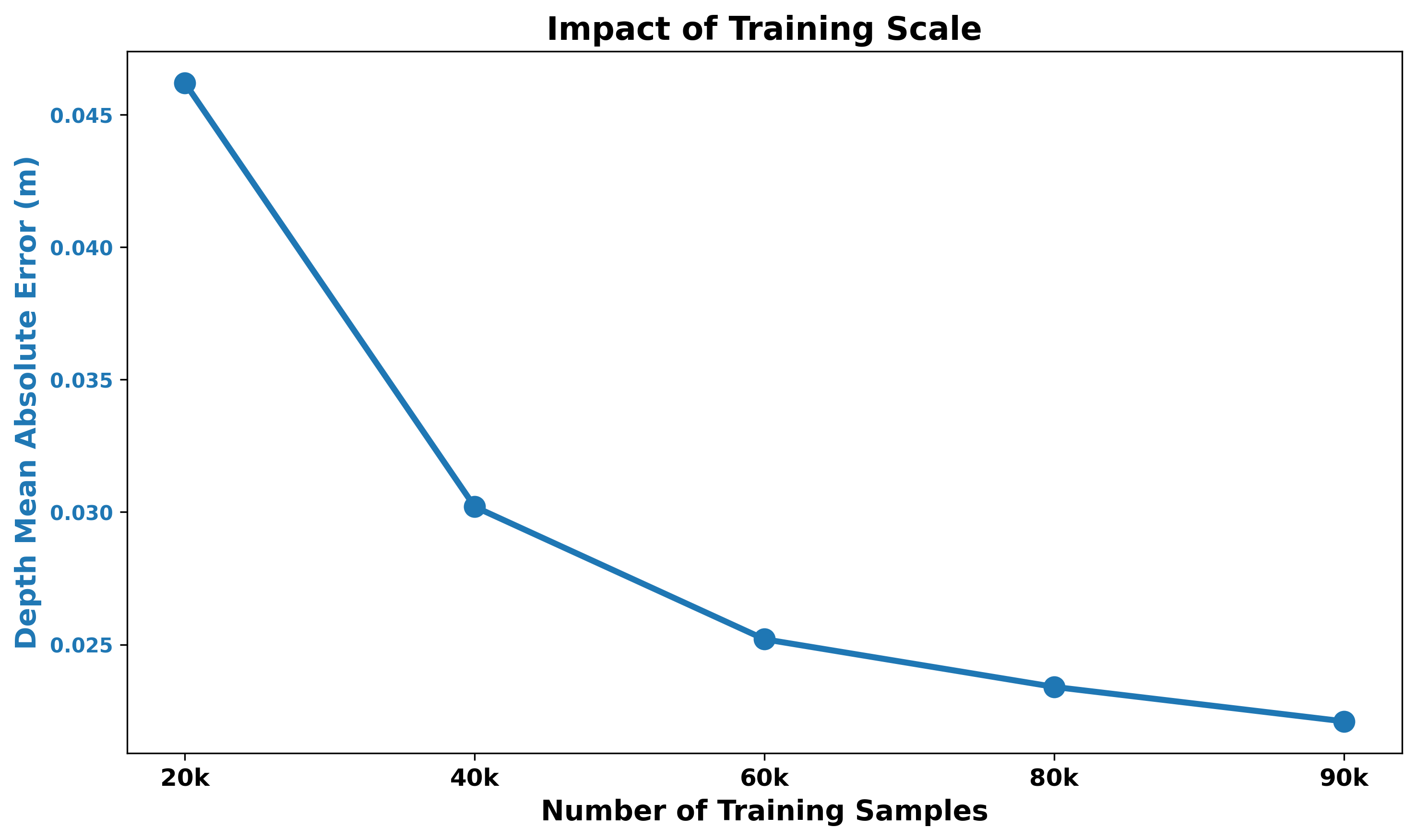}
    \caption{\textbf{Ablation: Training Scale.} Given the scale of the proposed SB3D Dataset, we ablate the impact of training data scale on performance, focusing on depth accuracy. We find that as the amount of training data increases from 20k to 90k samples, the depth error consistency is reduced. Interestingly, the depth error reduces below the limit set by the temporal resolution of the sensor -- meaning the model has learned precise correlations based on shape and appearance, rather than just relying on the timing information.
    }
    \label{fig:training_scale}
\end{figure}

\subsection{Impact of Training Scale}
One of the contributions of this work is a large-scale simulated transient dataset. To ablate the impact of its scale on learning from multi-bounce signals, we vary the amount of training data used for our depth model and study the impact on performance. While past work in deep learning has extensively shown that data scale is correlated with performance, we use this ablation to confirm that intuition in the case of single-photon lidar data. We find that as we increase the amount of training data from 20k to 90k samples, depth estimation accuracy continues to improve, as shown in \cref{fig:training_scale} of the main text.

\subsection{Out-of-Distribution Generalization}
We study the generalization capabilities of our model on out-of-distribution geometry. To do this, we rendered a new test dataset of multiplexed transients for 1,000 objects from Objaverse \cite{deitke2023objaverse} that don’t appear in our training dataset (animals, humans, etc.). Each scene contains a cuboid room with an object placed on the ground. We tested our depth and shadow estimation models on this dataset. Depth MAE was 0.0201 m and shadow MAE and IoU were 0.0541 and 92.4\% respectively, similar to previous in-distribution results (\cref{table:main_results}). We computed depth error only on object pixels (using an object mask), whereas shadow MAE/IoU were computed over all pixels.

\section{Implementation Details \& Architecture}
\label{sec:implementation}

\textbf{Pre-Processing.} Our method utilizes the raw multi-bounce lidar measurement as input, which, in our work had a shape $256 \times 256 \times 637$ in our main experiments and $256 \times 256 \times 375$ in our real-world experiments and noise ablation. We tried two methods for data normalization for the depth model: (1) reducing dynamic range by taking the log of each measurement followed by min-max normalization using the max intensity found over the entire dataset, and (2) min-max normalizing each histogram. While both were effective, we used (2) since it resulted in slightly better performance. For the shadow transient model, measured transients are min-max normalized and concatenated with the histogrammed predicted two-bounce ToF. For real-world data, the measurements were instead $256 \times 256 \times 375$ due to the differences in scene scale and temporal resolution.

\vspace{2mm}
\noindent
\textbf{Architecture.} Although the focus of our work is not on architecture, we investigated the efficacy of different architectures for the proposed tasks, including 2D U-Net \cite{ronneberger2015u}, ``2.5D'' U-Net (3D encoder and 2D decoder with learned projections in each skip connection), SwinIR \cite{liang2021swinir}, NLOST \cite{li2023nlost}, and NLOSFeatureEmbeddings \cite{chen2020learned}. We found that 2D U-Net and ``2.5D'' U-Net had the best performance -- with 2D U-Net training faster since larger batches could be fit on GPU. Thus, we used a modified 2D U-Net for all results. To accommodate the large size of our input, we added an initial feature extraction convolution to project 637 bins (or 375 for real-world data) to 128 channels before proceeding to the six U-Net encoder and decoder blocks.

\vspace{2mm}
\noindent
\textbf{Training.}
In simulation-based experiments, we trained three models: a ToF demultiplexing model, a shadow demultiplexing model, and a specular surface segmentation decoder (using the frozen ToF demultiplexing features). All models were trained for 200 epochs. The shadow model was trained with ground-truth two-bounce ToF data for the first 100 epochs and then with the noisier predicted two-bounce ToF data for the last 100 epochs. We found this curriculum learning strategy to be most effective to maximize accuracy. While the depth model and shadow model were both trained to generalize over scenes, the neural reconstruction method \cite{klinghoffer2024platonerf} was trained per scene.
\looseness-1

\vspace{2mm}
\noindent
\textbf{Implementation.} Our models are implemented in PyTorch \cite{paszke2019pytorch} and each trained on 8 NVIDIA H100 GPUs for around two days due to the size of the dataset used. We use the AdamW optimizer \cite{loshchilov2019decoupledweightdecayregularization} with an initial learning rate of $10^{-2}$ and weight decay of $10^{-3}$.

\section{Training / Eval / Test Splits for SB3D Dataset}
We train our models using the proposed dataset described in \cref{sec:dataset}. We use 90\% training split (87,688 samples), 3.9\% validation split (3,744 samples), and 6.1\% test split (6k samples). For 3D reconstruction, we train PlatoNeRF per scene using the predicted two-bounce ToF and shadows, as described in \cref{sec:method}. In all experiments we assume 25 illumination points in a grid pattern, unless stated otherwise (i.e., ablations on number of illumination points, noise ablation, and real-world experiments). %

\section{SB3D Dataset Rendering Details}
\begin{figure*}[t]
    \centering\includegraphics[width=0.95\textwidth]{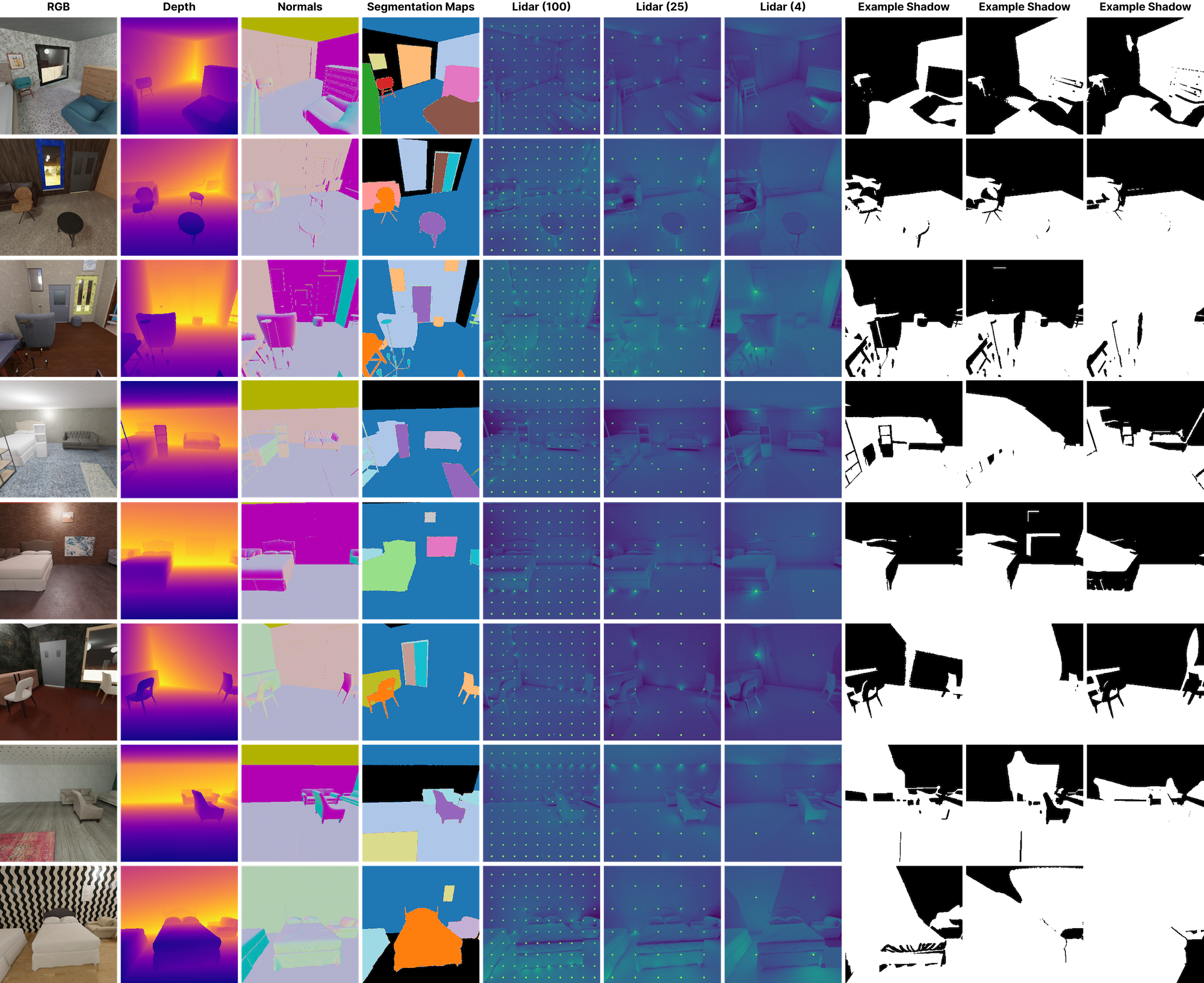}
    \caption{\textbf{SB3D Dataset (Extended)}. We provide additional examples from our proposed dataset. In total, the dataset contains 97,432 examples, each rendered from a different scene.
    }
    \label{fig:extended_dataset}
\end{figure*}

Meshes of indoor scenes are created from objects in the Amazon Berkeley Objects dataset \cite{collins2022abo} following the method and pipeline for procedural generation of realistic indoor scenes proposed for the creation of the Aria Synthetic Environments dataset \cite{avetisyan2024scenescript}. Objects are assembled to mimic realistic indoor environments. These configurations were shown to be sufficiently realistic for real-world generalization of models trained on rendered RGB in past work \cite{avetisyan2024scenescript}. Besides single-photon lidar, all renders are created with Blender. Single-photon lidar transients are created with the physically-based MitsubaToF renderer \cite{pediredla2019ellipsoidal}, which uses bidirectional path tracing with ellipsoidal connections to increase sampling efficiency. Due to the computational complexity of rendering single-photon lidar and the scale of the proposed dataset, rendering was parallelized over 1,000 CPU machines over around one week. All data is rendered at a resolution of $256 \times 256$ with a field of view of 90°. Multi-bounce lidar measurements are rendered with a temporal resolution of 128\,picoseconds or \textasciitilde0.0384\,meters. All bounces of light (1, 2, 3, and more) are rendered. To reduce rendering time, time gating is used when generating the transient data (all scenes have a minimum depth of no less than 0.5 m and a maximum depth of no more than 4.5 m). To ensure all two-bounce paths are recorded, pathlengths between 1 and 25.46\,meters are recorded, resulting in 637 bins per transient histogram. We set $n_{bounces}$ to -1 in MitsubaToF, meaning all bounces of light are rendered. Thus, the dataset can be used in future work that explores additional bounces. For any specular surface, we use the "roughconductor" BSDF in MitsubaToF and set alpha to 0.01. As a result, the lidar transients have either diffuse or specular surfaces (but not a gradient). We acknowledge this is a limitation of the dataset, as, in practice, many real-world objects may exhibit material properties with partial diffuse and specular components, however, in our real-world proof-of-concept experiments, we find this assumption is sufficient in demonstrating the potential for real-world generalization. For accessibility, the dataset is compressed to \textasciitilde5 TB for release. We provide additional examples from the proposed SB3D dataset in Fig. \ref{fig:extended_dataset}.

Although not used in this work, the proposed dataset contains over 30 instance label categories that can be used in future work on instance segmentation, including everyday objects, such as desks, chairs, books, beds, pillows, weights, and many more.

\section{PlatoNeRF Background}

Our method leverages PlatoNeRF \cite{klinghoffer2024platonerf}, a recent method for single-view 3D reconstruction from two-bounce transients. In contrast to our approach, PlatoNeRF assumes a laser is scanned over the scene sequentially, capturing separate transients for each laser spot. Each transient is preprocessed into two-bounce ToF and shadow masks, which are used to supervise the learned densities via volume rendering.

PlatoNeRF is trained in two stages. First, depth from the lidar to the scene is learned by tracing primary rays with volume rendering. Since ground truth depth is not directly available, predicted depth is used to compute two-bounce ToF based on known illumination point and laser location (i.e. tracing the distance from the laser to the illumination point, from the illumination point to the predicted scene point location, and from the predicted scene point location to the known sensor location). In the second stage, secondary rays are also traced. Secondary rays originate at the end of primary rays and go to each illumination point. Intuitively, the full path of the secondary ray is only traveled if the measured point is \textit{not} in shadow, else an object occludes the light from reaching the measured point. Thus, the secondary rays' transmittance values are supervised with the binary shadow masks extracted from the raw transient measurements.

Since PlatoNeRF uses two-bounce ToF and shadow masks to learn 3D scene geometry, the proposed pipeline naturally integrates with this approach for 3D reconstruction. Rather than computing these values from many non-multiplexed measurements, we instead use the values predicted by our models from multiplexed measurements, enabling single-shot 3D reconstruction.

\section{Real-World Experiments}

\begin{figure}
    \centering
    \includegraphics[width=1.0\linewidth]{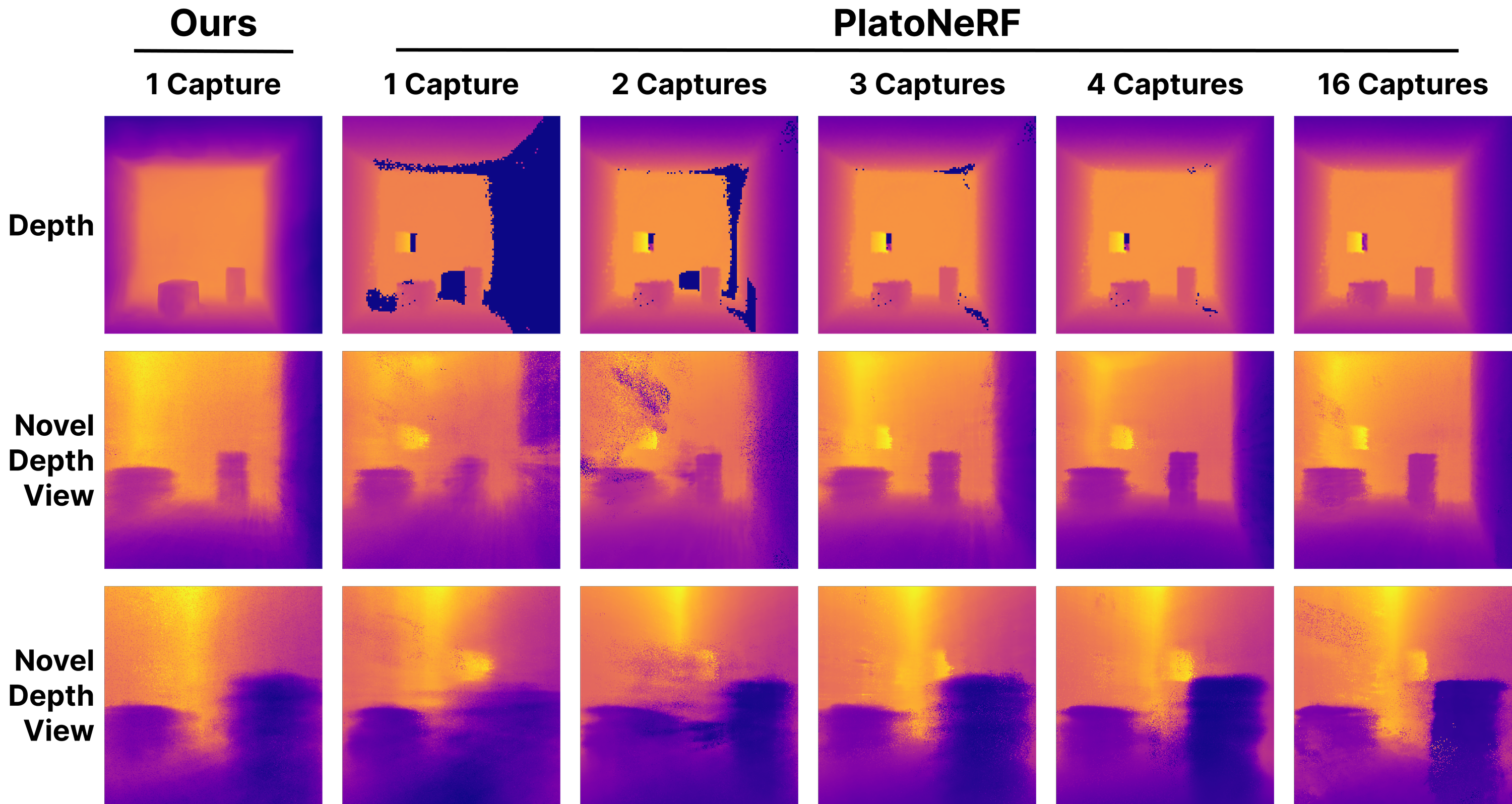}
    \caption{\textbf{PlatoNeRF Performance with More Captures.} While our method outperforms PlatoNeRF in the single capture setting, we compare to PlatoNeRF as more points are scanned and used in training. We find that, as expected, PlatoNeRF accuracy increases as more captures (with different illumination points) are added. Shoot-Bounce-3D remains competitive -- higher performance is especially noticeable in specular regions, though occluded regions have more floaters. 
    }
    \label{fig:platonerf}
\end{figure}

In this section, we elaborate on the details of our real-world dataset and proof-of-concept results.

\subsection{Model Training}
\label{sec:real_world_training}

For real-world validation, we retrain our models on a simulated dataset of scenes with a randomly placed cube, cylinder, and mirror in a room of varying scale. There are several reasons that motivate our use of a new training dataset for our real-world experiments. The original networks used (a) noiseless data, (b) scene scales too large for our galvo ranges and lab space (1–4 vs 0.4–1 m), (c) different camera settings (90° vs 45° FoV, 128 ps vs 32 ps bins), and (d) 25 illumination points (we used 16 to cut acquisition time). While differences in scene scale can be partially mitigated by rescaling our measurements, this process is approximate (e.g. the original number of bins $\times$ scene scale delta may not equal the target number of bins). Re-rendering the entire dataset to close these gaps would have been computationally expensive. Instead, we rendered a smaller dataset with the same scale, camera, and illumination as real, and added noise. Using this dataset allowed us to validate that transient demultiplexing is feasible under realistic signal and noise. The simulated scene is illuminated at 16 points simultaneously -- these points are in a grid pattern. When rendering, we randomly apply jitter to the camera origin and field of view for each scene. Each transient is rendered at 8 ps resolution and every four adjacent bins are summed to reduce the temporal resolution to 32 ps before training/inference. We choose 32 ps so that fine details can be resolved given the small scale of our scenes (the cylinder is one inch wide). Rendering with higher temporal resolution allows pulse shape and noise to be applied optionally before combining bins. The training dataset contained 10k samples, with an additional 2.5k for validation and 2.5k for test. We trained two sets of models -- one with noiseless data and one with added pulse shapes, noise, and timing jitter (see \cref{sec:noise} for details).

\subsection{Real-World Dataset}
\label{sec:real-world-dataset}

We capture a real-world dataset with scene geometry and sensor intrinsics/extrinsics that lie in distribution with the training data described above. We construct a room from diffuse white poster board and randomly place a foam cube and cylinder inside it, along with a mirror on the back wall. We illuminate each laser spot one at a time with a pulsed laser (Picoquant LDH-D Series) with 640 nm
wavelength and a two-axis scanning galvonometer (Thorlabs GVS412). For each laser spot, we then scan a single-pixel SPAD (MPD PDM Series) over a 46° field of view using a second two-axis scanning galvonometer. This procedure results in sixteen $256\times256$ transients captured at 8 ps resolution. We sum adjacent bins in each transient to reduce the temporal resolution to 32 ps and add the sixteen transients together to create a multiplexed measurement. The light-in-flight video is shown in the supplementary webpage. Ground truth depth is captured from 1-bounce light by converting the setup shown in Fig. \ref{fig:real-world} to be confocal. %

\begin{figure}
    \centering
    \includegraphics[width=0.85\linewidth]{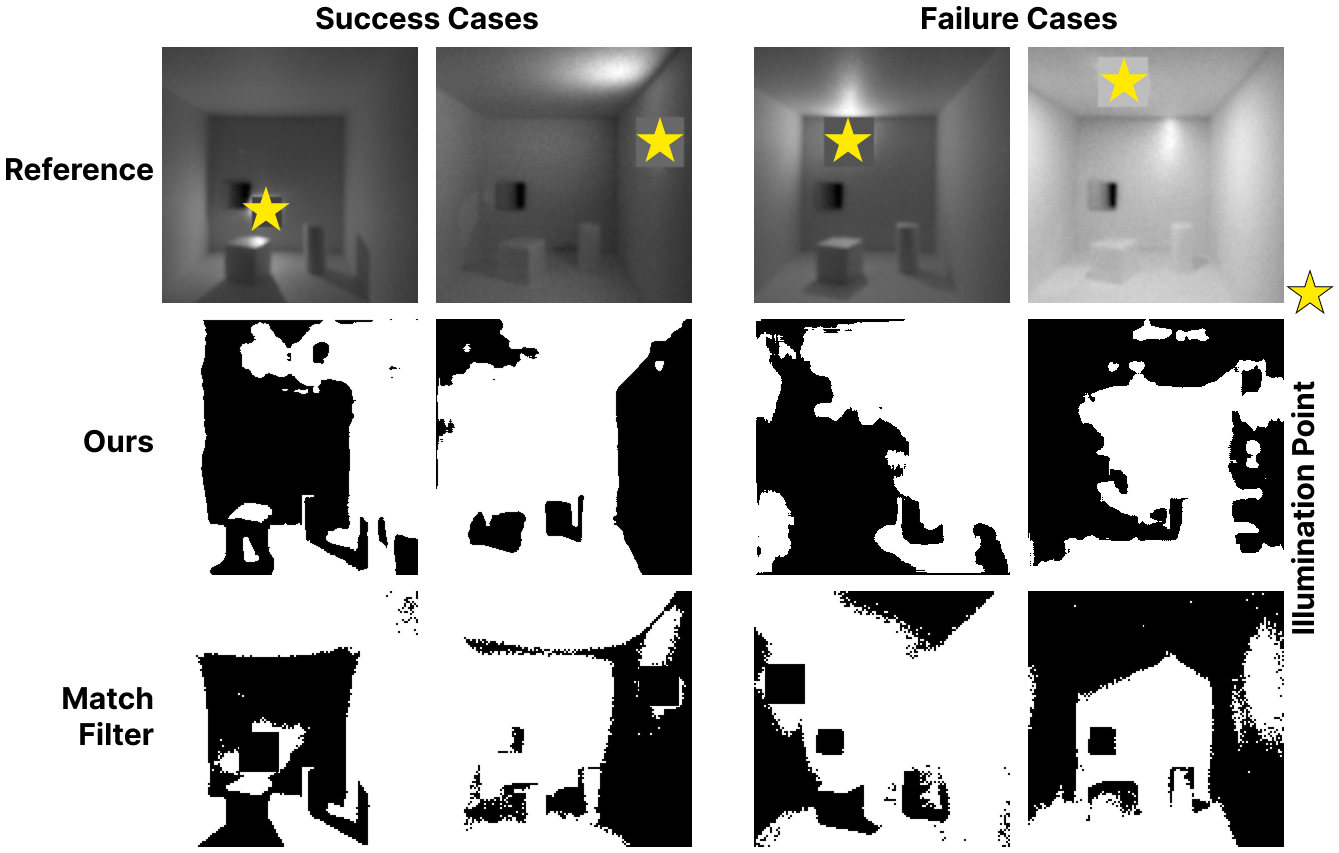}
    \caption{\textbf{Shadow Prediction Quality.} While our is able to produce shadows sufficient for 3D reconstruction, some predicted shadows have significant artifacts, as shown in the two examples on the right. Below our predicted shadows we show shadow quality when using a match filtr on the non-multiplexed measurement from the individual illumination point. 
    }
    \label{fig:real_shadows}
\end{figure}

\subsection{Results}

\begin{figure}
    \centering\includegraphics[width=0.9\linewidth]{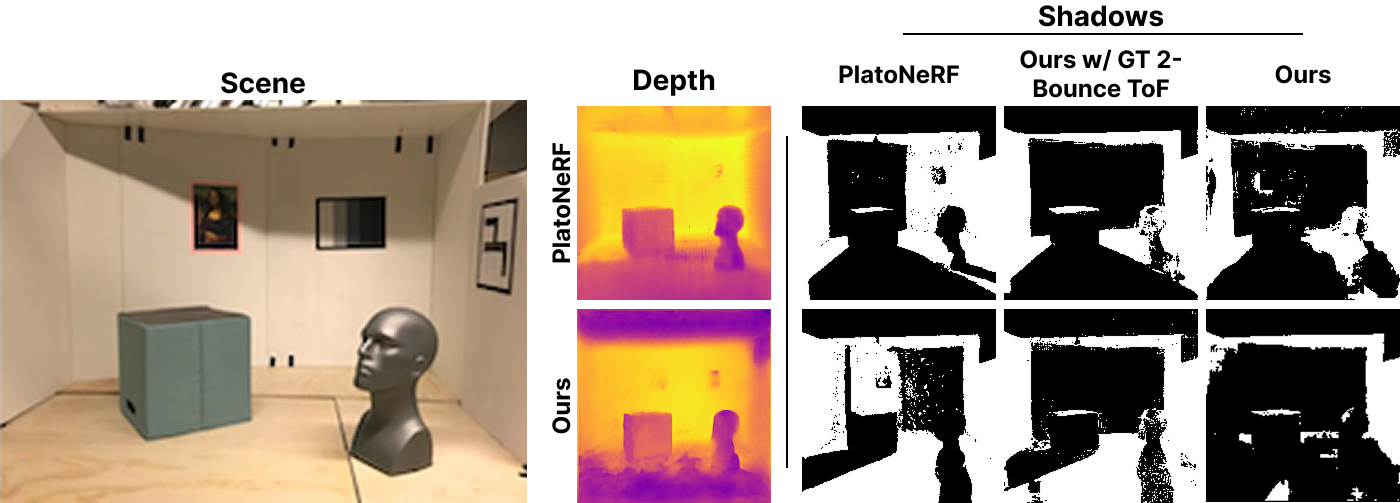}
    \caption{\textbf{Limitations in Generalizability.} 
    We test the generalizability of our models by testing them on an existing real-world dataset (from Bounce Flash Lidar \cite{henley2022bounce}) with different scene scale, geometry, and spatial/temporal sensor resolution than seen in training. Our models are able to predict reasonable depth despite these differences. However, while some structure is maintained in predicted shadows, there are noticeable artifacts, especially in the region near the mannequin's head. If we use ground truth 2-bounce ToF, along with raw transients, as input to our shadow model, instead of predicted 2-bounce ToF, shadow quality improves significantly. This improvement suggests that a limitation of our work is the propagation of errors from the depth estimation model to the shadow model.}
    \label{fig:bf_lidar_results}
\end{figure}

Results are shown in \cref{fig:real-world}. While we found our noiseless and noised models were both able to estimate reasonable depth, recovered shadows varied in quality, with some being highly accurate and others containing more artifacts, as shown in Fig. \ref{fig:real_shadows}. To find the best shadows, we performed a grid search over model (noiseless, noised), amount of noise to subtract, and maximum histogram intensity (for clipping). We also tried applying a low-pass filter to the data and performing peak finding to reconstruct histograms with Diracs, but found that neither improved performance. We used the four best shadows (via manual selection), along with the 2-bounce ToF, predicted by our model to train PlatoNeRF for 3D reconstruction. In Fig. \ref{fig:real-world}, we compare SB3D to PlatoNeRF trained when both are trained with only a single capture. Since PlatoNeRF is unable to handle multiplexed illumination, we instead train this PlatoNeRF model with a single illumination point. As shown in Fig. \ref{fig:platonerf}, as we increase the number of captures used to train PlatoNeRF, its performance improves. SB3D outperforms PlatoNeRF in the single capture setting and remains competitive with PlatoNeRF even when PlatoNeRF is trained with 16 captures. Training PlatoNeRF with 16 captures (by scanning a laser over different illumination points) serves as an upper bounce on 3D reconstruction. With 16 captures, PlatoNeRF exhibits slightly fewer floaters/artifacts in occluded regions than SB3D, but SB3D exhibits better performance in areas with specular objects due to its use of a data prior.

\paragraph{Limitations \& Opportunities} Our real-world results demonstrate feasibility that the ideas proposed in this work can extend to real-world settings, In this section, we investigate generalizability and limitations of our model.

We test our models' ability to generalize to another real-world dataset, from BF Lidar \cite{henley2022bounce}. This test is challenging because the scene in this dataset has different scale and geometry (e.g. a mannequins head) than the scenes in our training dataset. In addition, the multiplexed measurement from BF Lidar has different spatial and temporal resolution than our models were trained with. Specifically, our models from Sec. \ref{sec:real_world_training} were trained with $256\times256$ spatial resolution and 32 ps temporal resolution, whereas the BF Lidar data is $200\times200$ spatial resolution with 128 ps temporal resolution. In addition, the BF Lidar scene is illuminated at 16 random points, rather than in a grid pattern. To account for this, we retrain the models described in Sec. \ref{sec:real_world_training} with random illumination points for every training sample, testing whether our model can not only generalize to a real-world measurement with different geometry and resolution, but also generalize to random illumination patterns.

To test our models on the BF Lidar dataset, we zero-pad the measurements and rescale the detected two-bounce peaks bins based on the difference in scene scale between training and test. Results are shown in Fig. \ref{fig:bf_lidar_results}. Despite the significant domain gaps, our model is capable of predicting reasonable depth, albeit with artifacts. The predicted shadows contain accurate regions, but also regions with significant artifacts. To understand the cause of these artifacts, we tried using ground-truth 2-bounce ToF, rather than predictions, -- along with raw lidar transients -- as input to our shadow estimation model. This experiment resulted in significant improvements in shadow quality. This finding suggests that the shadow models are also able to generalize to different geometries and sensor resolutions if given accurate 2-bounce ToF, but errors in depth estimation propagate and can significantly impact the shadow model.

Future work may explore different types of noise to add to the two-bounce ToF during training to improve robustness or ways to unify the first two stages of our approach. Other improvements may come from incorporating real-world data into training and investigating other ways to mitigate the sim-to-real gap.

\end{document}